\documentclass[letterpaper, 10 pt, conference]{ieeeconf}  % Comment this line out if you need a4paper

\IEEEoverridecommandlockouts                              % This command is only needed if
\usepackage{amssymb}
\usepackage{graphicx}
\usepackage{algorithm}
\usepackage{hyperref} % 应该先加载 hyperref
\usepackage{cleveref}
\usepackage{pifont}
\usepackage{multirow}
\usepackage{booktabs}
\usepackage{wrapfig}
\usepackage{caption}
\usepackage[dvipsnames, table]{xcolor}  % 支持更多颜色
\usepackage{mwe}
\usepackage{xspace}
\usepackage{footmisc} % 可选，用于更多脚注控制
\usepackage{listings}
\usepackage[normalem]{ulem}
\useunder{\uline}{\ul}{}

\definecolor{lightgray}{gray}{0.95}

\lstdefinelanguage{Prompt}{
    morekeywords={ROOM_PROMPT},
    sensitive=false,
    morecomment=[l]{\#},
    morestring=[b]",
}

\usepackage{titlesec}

\usepackage{algorithm}
\usepackage{algpseudocode}
\algrenewcommand\algorithmicindent{0.8em} % 设置缩进为 1 个字符宽
\usepackage{wrapfig}

\newcommand{\code}[1]{{\texttt{#1}}}

\newcommand{\cmark}{\textcolor{ForestGreen}{\ding{51}}}       % 绿色✓
\newcommand{\xmark}{\textcolor{Maroon}{\ding{55}}}           % 红色×

\newcommand{\icra}[1]{\textcolor{black}{#1}}
\newcommand{\zongtai}[1]{\textcolor{black}{#1}}
\newcommand{\haokun}[1]{\textcolor{black}{#1}}

\newcommand{\ourmethod}{\texttt{STRIVE}\xspace}
\newcommand{\ourmethodplain}{{STRIVE}\xspace}

\crefname{equation}{Eq.}{Eqs.} \crefname{section}{Sec.}{Secs.} \crefname{subsection}{Sec.}{Secs.}
\crefname{figure}{Fig.}{Figs.} \crefname{table}{Tab.}{Tabs.} \crefname{algorithm}{Alg.}{Algs.}
\crefname{theorem}{Thm.}{Thms.} \crefname{lemma}{Lem.}{Lems.} \crefname{corollary}{Cor.}{Cors.}
\crefname{proposition}{Prop.}{Props.} \crefname{definition}{Def.}{Defs.}
\crefname{remark}{Rem.}{Rems.} \crefname{example}{Ex.}{Exs.} \crefname{appendix}{App.}{Apps.}
\crefname{line}{Line}{Lines} \crefname{footnote}{Fn.}{Fns.} \crefname{enumi}{Item}{Items}

\title{\LARGE \bf
\ourmethodplain: Structured Representation Integrating \\ VLM Reasoning for Efficient Object Navigation
}

\author{
	\textbf{Haokun Zhu}$^{1*}$,  
	\textbf{Zongtai Li}$^{1*}$, 
	\textbf{Zhixuan Liu}$^{1}$, \\
	\textbf{Wenshan Wang}$^{1}$,
	\textbf{Ji Zhang}$^{1}$, 
	\textbf{Jonathan Francis}$^{1,2}$, 
	\textbf{Jean Oh}$^{1}$ \\
	$^{1}$Carnegie Mellon University \quad 
	$^{2}$Bosch Center for AI \\
}

\begin{document}

% \maketitle
% \thispagestyle{empty}
% \pagestyle{empty}

\twocolumn[{%
\renewcommand\twocolumn[1][]{#1}%
\maketitle
\begin{center}
    \centering \small
    \vspace{-15pt}
    \captionsetup{type=figure}
    \includegraphics[width=0.96\linewidth]{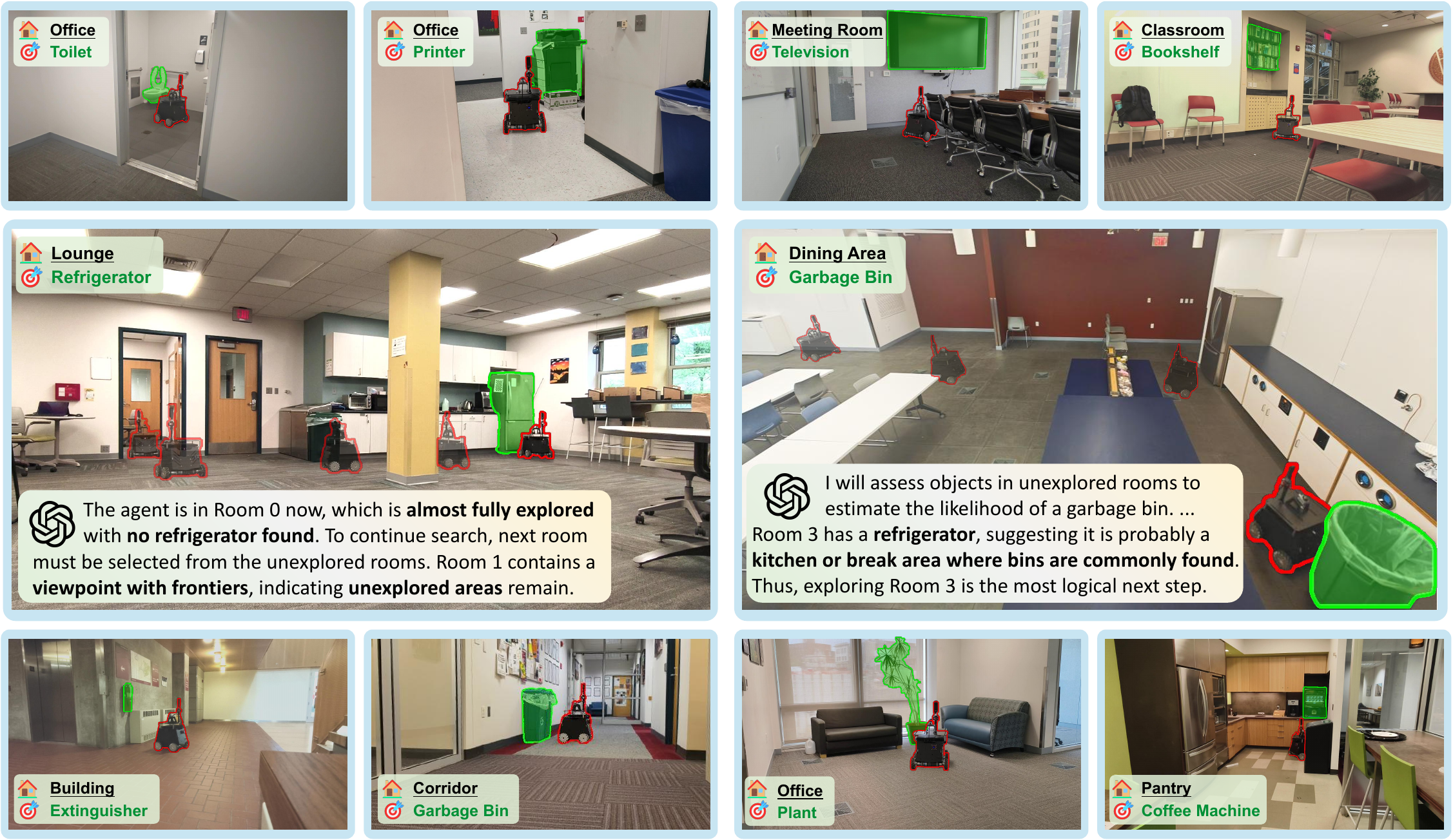}
    \captionof{figure}{{\ourmethodplain can conduct zero-shot object navigation in diverse and complex real-world environments by leveraging our novel multi-layer representation and efficient two-stage navigation policy.}}%
    % \vspace{-5pt}
    \label{fig:teaser}
\end{center}%
}]
\let\thefootnote\relax\footnotetext{$^{*}$Equal contribution.}

%%%%%%%%%%%%%%%%%%%%%%%%%%%%%%%%%%%%%%%%%%%%%%%%%%%%%%%%%%%%%%%%%%%%%%%%%%%%%%%%
\begin{abstract}
    Vision-Language Models (VLMs) have been increasingly integrated into object navigation tasks for their rich prior knowledge and strong reasoning abilities.  However, applying VLMs to navigation presents two key challenges: effectively \icra{parsing and structuring} complex environment information and determining \textit{when and how} to query VLMs. Insufficient environment understanding and over-reliance on VLMs (e.g. querying at every step) can easily lead to unnecessary backtracking and reduced navigation efficiency, especially in large continuous environments.
    To address these challenges, we propose a novel framework that \zongtai{incrementally} constructs a multi-layer environment representation consisting of viewpoints, object nodes, and room nodes during navigation. Viewpoints and object nodes facilitate intra-room exploration and accurate target localization, while room nodes support efficient inter-room planning.
    Building on this structured representation, we propose a novel two-stage navigation policy, integrating high-level planning guided by VLM reasoning with low-level VLM-assisted exploration to efficiently \icra{and reliably} locate a goal object.
    We evaluated our approach on three simulated benchmarks (HM3D \icra{v1\&v2}, RoboTHOR, and MP3D), and achieved state-of-the-art performance on both the success rate (SR\% $\mathord{\uparrow}\, 13.1\%$
    ) and navigation efficiency (SPL\% $\mathord{\uparrow}\, 6.2\%$
    ). We further validate our method on a real robot platform, demonstrating strong robustness across \zongtai{120 episodes} in 10 different indoor environments.
    % Videos are available \href{https://anonymous.4open.science/w/STRIVE-459/}{https://anonymous.4open.science/w/STRIVE-459/}.
    Project page is available at \href{https://zwandering.github.io/STRIVE.github.io/}{\textit{https://zwandering.github.io/STRIVE.github.io/}}.
\end{abstract}

%%%%%%%%%%%%%%%%%%%%%%%%%%%%%%%%%%%%%%%%%%%%%%%%%%%%%%%%%%%%%%%%%%%%%%%%%%%%%%%%
\section{Introduction}
\label{sec:intro}
% We propose STRIVE: Structured Representation Integrating VLM Reasoning for Efficient Object Navigation.

% 第一段阐述object navigation task的重要性以及难点，证明其是一个值得做并且unsolved的问题
% 第二段阐述因为以上难点，recent 方法采用vlm或者llm的commensense reasoning来辅助navigation，但是有以下两个问题：1）输入问题，局限于local的输入，忽略历史信息，或者缺乏对历史信息的静态，层次化建模。因此VLM推理时缺少系统的上下文支撑。 2）没有使用结构化的历史轨迹或环境建模来约束或引导VLM的推理。Baseline在每一步决策时，让VLM直接在所有可达viewpoints中选择, 但由于vlm缺乏3D理解能力，无法结合空间layout和历史轨迹综合评估每个viewpoint，导致对于每个viewpoint的评估基本上都只是是基于局部的semantic information，容易做出冗余的决策（例如走回头路、重复探索已经经过的区域），导致navigation efficiency的下降。
% 第三段阐述我们的方法，我们提出来STRIVE，一个framework来incrementally构建一个structured representation，来辅助vlm reasoning。这个representation有三个层次：object nodes, viewpoints and room nodes。object nodes表示环境中所有观察到的objects，设计用来提供全面的semantic information和做target localization。viewpoints离散化环境为一组locations，设计用来做高效的intra-room navigation。room nodes进一步将环境分割为不同的房间，room node的存在使得VLMs可以根据不同room的layout和semantic信息，在room level进行推理。这个层次化的representation允许对环境进行更全面的总结，充分利用VLM的推理能力，在复杂的环境中做出更有效的决策。
% 并且在这个representation的基础上，设计了一个高效的room-to-room的navigation policy，我们通过对于环境的层次化建模来引导VLMs的推理。具体来说，我们只让VLM根据每个room的layout和semantic信息来选择下一个要探索的room，而不是在每一步都让VLM直接在所有可达viewpoints中选择。在VLM选择的房间内部，我们使用传统的frontier-based算法探索viewpoints。这样有效的规避了VLMs对3D空间理解能力不足的问题，防止了VLMs在每一步都要做出冗余的action，提高了navigation的效率。我们还设计了early stopping和time-penalized relocation机制来进一步提高效率。
% 我们的方法在多个数据集上都取得了state-of-the-art的结果，并且在真机上也表现出了良好的性能。
Object navigation is a fundamental task in robotics, where an agent must locate an instance of a given object category in unknown environments. This task is particularly challenging, as it requires the agent to understand complex visual information, reason about spatial relationships, and make decisions based on both current and past observations.

Advances in Vision-Language Models (VLMs)~\cite{radford2021learning,wu2024gpt4visionhumanalignedevaluatortextto3d,team2023gemini} have demonstrated strong capabilities in contextual visual understanding and common-sense reasoning. Building on this, recent works~\cite{yin2024sg,longinstructnav,wu2024voronav,saxena2024grapheqa,loo2024open} have integrated VLMs into object navigation tasks, utilizing their rich prior knowledge, visual understanding, and commonsense reasoning abilities to guide navigation.
% Recent advances in Vision-Language Models (VLMs) have shown strong cross-task performance. Building on this, recent robotics works have leveraged VLMs in object navigation tasks, relying on the models' rich prior knowledge, in-context learning capabilities, and commonsense reasoning abilities for guiding navigation and goal satisfaction \cite{hu2023toward, yenamandra2024towards, rana2023sayplan}.
However, existing approaches often face two significant challenges:
% in using VLMs for object navigation tasks.
First, the input to VLMs typically lacks a structured representation of the environment and is often restricted to local observations. Without a coherent global view that integrates both current and previous observations, VLMs struggle to reason effectively about the environment and fail to make reasonable navigation decisions.
Second, existing methods~\cite{wu2024voronav, loo2024open} typically rely on VLMs to select among all frontier viewpoints at each step, without utilizing navigation progress or environment layouts to effectively guide VLMs' reasoning process. Besides, due to VLMs’ limited understanding of 3D spatial information~\cite{zhang2024vision, chen2024spatialvlm, qi2025gpt4scene}, they cannot jointly reason about the spatial relationships and the navigation history when evaluating each viewpoint. As a result, their evaluation of viewpoints is largely based on viewpoints' local semantic information, which often leads to redundant navigation behaviors such as backtracking or repeated exploration. 

To address these challenges, we propose \ourmethod (\textbf{ST}ructured \textbf{R}epresentation \textbf{I}ntegrating \textbf{V}LM Reasoning for \textbf{E}fficient Object Navigation), a novel framework that incrementally learns a structured representation of the environment and utilizes VLMs' reasoning abilities to guide the navigation.
This representation consists of 3 layers: object nodes, viewpoint nodes, and room nodes. Object nodes represent all observed objects, 
% \zongtai{encapsulate rich semantic information} 
\haokun{provide rich semantic information}
%provide comprehensive semantic information
about the environment and assist in target localization;
Viewpoint nodes discretize the environment into a set of key locations, \zongtai{facilitating structured and efficient intra-room exploration}; %enabling efficient intra-room exploration;
Room nodes further segment the environment into distinct rooms and facilitate room-level reasoning by the VLM.
This multi-layer representation enables a more comprehensive understanding of the environment, allowing VLM to better utilize their reasoning abilities for more effective decision-making.
% in complex environments.
Furthermore, we design an efficient two-stage navigation policy based on this representation, combining high-level planning guided by the VLM's reasoning and VLM-assisted low-level exploration.
Specifically, for the high-level planning, instead of making step-by-step decisions among all viewpoint nodes, the VLM selects the next room to explore based on the spatial layout and semantic information of each room. For low-level exploration within rooms, we employ a traditional frontier-based algorithm for efficient exploration, while leveraging the VLM to decide whether continued exploration of the current room is \zongtai{necessary} %worthwhile
. Making high-level planning on rooms effectively mitigates the issue of VLMs' insufficient 3D spatial understanding and prevents redundant actions, thereby enhancing navigation efficiency.

We evaluate our method on \icra{four} widely-used simulated benchmarks: \icra{HM3D-v1}~\cite{habitatchallenge2022}, \icra{HM3D-v2}~\cite{habitatchallenge2023}, RoboTHOR~\cite{RoboTHOR}, and MP3D~\cite{Matterport3D}. \ourmethod achieves state-of-the-art (SOTA) results, significantly outperforming 12 existing methods in both Success Rate (SR) and navigation efficiency, measured by Success weighted by Path Length (SPL). This highlights the effectiveness of our proposed multi-layer representation and the VLM-guided reasoning policy in improving object navigation. Specifically, \ourmethod achieves \icra{62.9\% SR and 34.2\% SPL on HM3D-v1, 79.6\% SR and 38.7\% SPL on HM3d-v2}, 68.1\% SR and 36.3\% SPL on RoboTHOR, and 52.3\% SR and 23.1\% SPL on MP3D. Besides, we also conduct 15 real-world experiments across 10 different indoor environments on a Mecanum wheel platform~\cite{autonomy_stack_mecanum_wheel}, demonstrating the effectiveness and robustness of our method in real-world scenarios.

\section{Related Works}
\label{sec:related_works}
\begin{figure*}[t]
    \centering
    \vspace{5pt}
    \includegraphics[width=\linewidth]{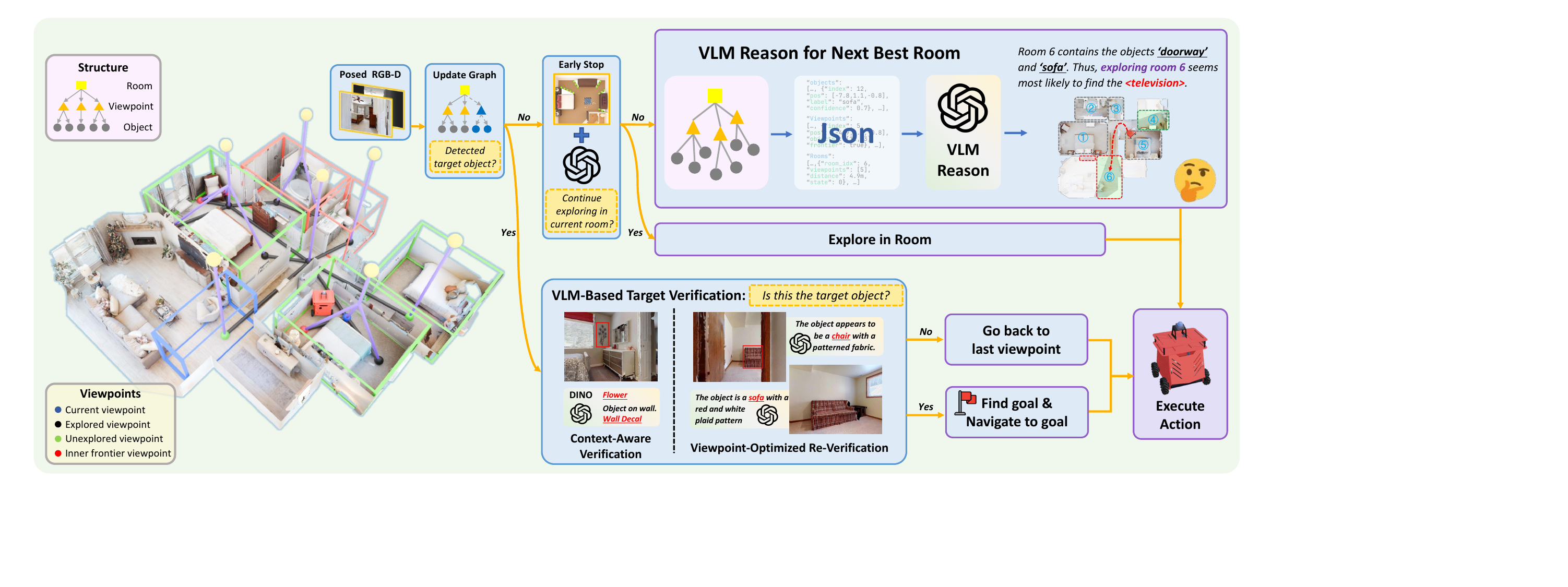}
    \caption{\textbf{Overview of \ourmethodplain}\textbf{.} We construct a multi-layer representation $\mathcal{R}$ (\cref{sec:graph-construction}) on-the-fly, consisting of object, viewpoint, and room nodes, which serves as a structured input for VLM. Based on $\mathcal{R}$, we introduce a two-stage navigation policy, where the VLM reasons and plans at room-level (\cref{sec:room}), while the agent explores in room at the viewpoint-level using a VLM-assisted frontier-based navigation strategy (\cref{sec:explore-in-room}) and %. The agent also performs
    VLM-based target verification (\cref{sec:object-found}).}
    \vspace{-5pt}
    \label{fig:pipeline}
\end{figure*}

\noindent\textbf{Object Navigation.} %
Existing object navigation methods are typically categorized into end-to-end learning approaches and modular approaches. End-to-end methods~\cite{dang2023search, wijmans2019dd, ye2021auxiliary, ramrakhya2022habitat, ramrakhya2023pirlnav, mousavian2019visual, yang2018visual} use reinforcement learning to directly map observations to actions, but often suffer from low sample efficiency and poor generalization. In contrast, modular methods~\cite{chaplot2020object, ramakrishnan2022poni, zhang20233dawareobjectgoalnavigation, Yu_2023, zhou2023escexplorationsoftcommonsense, chaplot2020object, yin2024sg, gu2025doraemon} decompose navigation into steps such as mapping, planning, and action execution, and often build semantic maps in bird's-eye view or 3D space to facilitate more interpretable and scalable navigation behavior.
With the emergence of foundation models~\cite{radford2021learning}, object navigation has advanced towards zero-shot, open-vocabulary setting~\cite{Yu_2023, zhou2023escexplorationsoftcommonsense, yokoyama2024vlfm, yin2024sg, gu2025doraemon}. %VLFM~\cite{yokoyama2024vlfm} aligns object goals with CLIP embeddings, while
We also leverage VLM's reasoning abilities to improve zero-shot object navigation.
but we employ a novel representation and a two-stage policy, enabling more efficient and effective VLM guidance.
%but we query the VLM to guide the navigation process more efficiently and effectively by introducing a novel representation and a two-stage policy.
% However, we argue that {over-reliance on LLMs} for decision-making in large, continuous environments can lead to {inefficient navigation}. To address this, we propose to combine VLM-based high-level planning with traditional exploration strategies, thereby leveraging the reasoning strength of LLMs while ensuring efficient and robust navigation behavior.
% With the rise of language models and multi-modal foundation models like CLIP \cite{radford2021learningtransferablevisualmodels}, Object Navigation has advanced towards zero-shot open-vocabulary navigation. For instance, L3MVN \cite{Yu_2023} and ESC \cite{zhou2023escexplorationsoftcommonsense} use LLMs for decision-making over candidate goals, SG-Nav \cite{yin2024sg} constructs 3D scene graphs and prompts LLMs with structural relationships, and CogNav \cite{cao2024cognav} explicitly modeling the cognitive process of object navigation utlizing LLM reasoning.
% Unlike prior work, we argue that {over-reliance on LLMs} for decision-making in large, continuous environments can lead to {inefficiency and reduced accuracy}. To address this, we propose to combine LLM-based high-level planning with traditional exploration strategies, thereby leveraging the reasoning strength of LLMs while ensuring efficient and robust navigation behavior.

\noindent\textbf{VLM-guided Navigation.} %
%\label{sec:related_works_llm_guided_navigation}
% copy from cognav, instructnav
% 然而，由于VLM缺乏3D空间感知能力，过分依赖VLM进行导航可能会导致低效的导航行为（频繁的backtrack）。为了解决这个问题，我们提出来一个两阶段navigation policy，combining VLM-based high-level planning with traditional exploration strategies, thereby leveraging the reasoning strength of VLMs while ensuring efficient and robust navigation behavior.
With internet-scale training data, Vision-Language Models (VLMs)~\cite{li2022blipbootstrappinglanguageimagepretraining, radford2021learning, wu2024gpt4visionhumanalignedevaluatortextto3d, team2023gemini} have shown strong commen-sense reasoning abilities and have been widely applied in Object Navigation tasks to guide the decision-making process. For example, InstructNav~\cite{longinstructnav} leverages multi-sourced value maps to model key navigation elements. SG-Nav~\cite{yin2024sg} constructs 3D scene graphs and prompts LLMs with structural relationships, while \icra{DORAEMON~\cite{gu2025doraemon} combines RAG-VLM for understanding ontological tasks and Policy-VLM for enhanced policy planning} However, due to VLMs' limited 3D spatial understanding ability~\cite{zhang2024vision, chen2024spatialvlm, qi2025gpt4scene}, over-reliance on VLMs for navigation can lead to inefficient behavior, such as frequent backtracking. To address this, we propose a two-stage navigation policy that combines VLM-guided high-level planning with VLM-assisted frontier-based low-level exploration strategies, leveraging the reasoning strength of VLMs while ensuring efficient and robust navigation behavior.

\noindent\textbf{Scene Representation for Indoor Navigation.} %
%\label{sec:related_works_scene_representation}
% copy from VoroNav: Voronoi-based Zero-shot Object Navigation with Large Language Model or somewhere else
% In navigation tasks, scene representation plays a crucial role in transforming raw observations into structured information that can be efficiently utilized for decision-making. Frontier-based methods \cite{ramakrishnan2022poni, chen2023not, gadre2023cows, gervet2023navigating} store frontiers in grid map, integrate semantic information into the map, and update in real-time.
Scene representation is crucial for transforming raw observations into structured information for decision-making in navigation tasks. Frontier-based methods~\cite{ramakrishnan2022poni, chen2023not, gadre2023cows, gervet2023navigating} record frontiers on a grid map and integrate semantic information to guide exploration.
In contrast, graph-based methods represent the environment as structured scene graphs to support navigation. Prior works~\cite{yin2024sg,loo2024open} use scene graphs to summarize semantic information and let VLMs to select among frontier locations. Others~\cite{an2024etpnav,wu2024voronav} explicitly construct viewpoints in the scene graph, enabling VLMs to reason over the graph and choose among viewpoints to guide navigation.
% We also consturct viewpoint nodes to discretize the environment. Unlike traditional scene graphs where viewpoints are often derived from static Voronoi graphs, our method segments the environment into local regions based on current exploration status and build  viewpoints according to the regions using a frontier clustering algorithm that prioritizes exploration efficiency. These structured viewpoints, together with the multi-layer representation, enable VLMs to reason more effectively about both the spatial structure and semantic content of the environment.
Unlike traditional scene graphs~\cite{hughes2022hydra,wu2024voronav}, where viewpoints are typically derived from Voronoi partitions, we discretize the environment into semantically meaningful regions to select viewpoint nodes. As the middle layer, these viewpoints bridge the spatial structure (room nodes) and semantic content (object nodes), forming a structured representation facilitating VLM reasoning.

% In contrast, graph-based methods \cite{krantz2020beyond, an2024etpnav, wu2024voronav, yin2024sg} extract a topological representation of the entire environment, where viewpoints are distributed across the space to represent distinct locations, gathering surrounding geometric and semantic information. This approach provides a clearer understanding of the environment’s layout, enhancing decision-making during the navigation process.
\section{Method}
\label{sec:method}
\textbf{Task Definition:}
In Object Navigation, the agent is required to find an instance of a given object category (e.g. Find the \textit{bed}.) in an unknown environment. At each time step $t$, the agent receives a posed RGB-D observation $\mathbf{O}_t = \{I_t, D_t, P_t=\langle \mathbf{p}_{t}, \mathbf{R}_t \rangle\}$, where $I_t$ is the RGB image, $D_t$ is the depth map, and $P_t$ is the camera pose.
The navigation policy then predicts an action $a_t \in \{\texttt{move\_forward, turn\_left, turn\_right, stop} \}$. The task is considered successful if the agent stops within $d_s$ meters of the target object in less than $T$ steps.

\textbf{Overview:}
\cref{fig:pipeline} provides an overview of \ourmethod.
%We construct a hierarchical representation of the environment, comprising three layers: the object nodes, the viewpoints, and the room nodes.
% We construct a multi-layer representation of the environment, and the graph construction process is detailed in \cref{sec:graph-construction}.
% Unlike previous methods that rely on a dense and continuous action space of the entire scene, our approach discretizes the scene using viewpoint nodes, enabling the creation of a sparse and discrete action space where the robot operates only on these nodes.
% \zongtai{Previous methods allow the agent to freely select any location within the scene, requiring the language model to reason over a large and continuous action space. In contrast, our approach discretizes the scene using viewpoint nodes, enabling a sparse and discrete action space.}
% We also design an efficient two-stage navigation policy that leverages the structured representation to guide VLM reasoning at the room level. The details of this policy are presented in \cref{sec:object-navigation-policy}.
% \zongtai{a framework that constructs a multi-layer environment representation and performs object navigation through a novel two-stage navigation policy.}
\ourmethod enables the VLM to reason at the room level while \zongtai{guiding the agent's in-room exploration} at the viewpoint level. We describe the representation construction in \cref{sec:graph-construction} and the two-stage navigation policy in \cref{sec:object-navigation-policy}.
%The representation construction process is detailed in \cref{sec:graph-construction} and the two-stage navigation policy is presented in \cref{sec:object-navigation-policy}.

% 这个想放的话感觉放到 viewpoints 那边或者 3.1 那边，overview有点太长了
% Previous methods allow the agent to select any location or all unvisited viewpoints in the scene, requiring the language model to reason over a large and continuous action space. In contrast, our approach discretizes the scene using viewpoints and room nodes, enabling a sparse and discrete action space for VLMs to reason over and for the robot to navigate on.
% \todo{Figure xxx provide an overview of our representation and navigaiton policy. We first build xxx, object layer, viewpoint layer, room layer. 其中viewpoint layer离散化了整个场景，room layer提供了一个全局的导航策略。Based on that, we develop XXX}

\subsection{Multi-layer Environment Representation}
\label{sec:graph-construction}

We propose a framework that models the environment using a three-layer graph representation $\mathcal{R}$, where each layer corresponds to a specific type of node: object nodes $V^{obj} = \{v_i^{obj}\}$, viewpoint nodes $V^{vp} = \{v_i^{vp}\}$, and room nodes $V^{room} = \{v_i^{room}\}$. Edges encode spatial and semantic relationships across nodes. We elaborate on the graph construction process in following sections.
% \zongtai{remove this? Each layer serves a specific purpose in the navigation process. We will elaborate on the construction of each layer in the following sections.}
\subsubsection{Viewpoint Nodes}
\label{sec:viewpoint-node}
Inspired by~\cite{yang2021graph}, we construct a skeleton graph to discretize the environment. The graph is incrementally built as the agent navigates—each time the agent reaches a new viewpoint, it updates the graph. We define a coverage range $\zeta_{cover}$ as the maximum distance within which \zongtai{objects are assumed to be reliably detected}. %semantic information is associated with the center viewpoint.
\zongtai{Each viewpoint node thus controls a circular region of radius $\zeta_{cover}$.}
% Each viewpoint node thus controls a local region determined by $\zeta_{cover}$.
Edges between viewpoint nodes indicate direct traversability. The maximum sensor range $\zeta_{max}$ denotes the effective measurable distance of the depth camera.

\begin{figure}
    \centering
    \includegraphics[width=0.8\linewidth]{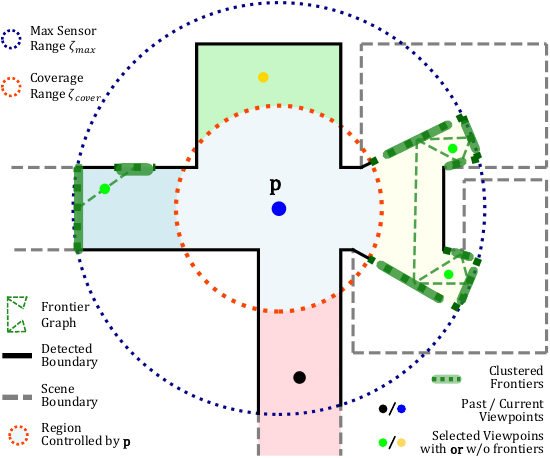}
    \caption{\textbf{Visualization of the viewpoint selection algorithm.} \textcolor{Green}{\textit{Green}} and \textcolor{Dandelion}{\textit{yellow}} nodes are the selected viewpoints.}
    \label{fig:viewpoint}
    \vspace{-18pt}
\end{figure}

% \begin{wrapfigure}
%     {r}{0.42\textwidth}
%     \vspace{-13pt}
%     \centering
%     \includegraphics[width=0.42\textwidth]{figures/viewpoint_5.pdf}
%     \vspace{-12pt}
%     \caption{\textbf{Visualization of the viewpoint selection algorithm.} \textcolor{Green}{\textit{Green}} and \textcolor{Dandelion}{\textit{yellow}} nodes are the selected viewpoints.}
%     \label{fig:viewpoint}
%     \vspace{-10pt}
% \end{wrapfigure}

\begin{figure*}[t]
    \centering
    \vspace{5pt}
    \includegraphics[width=0.85\linewidth]{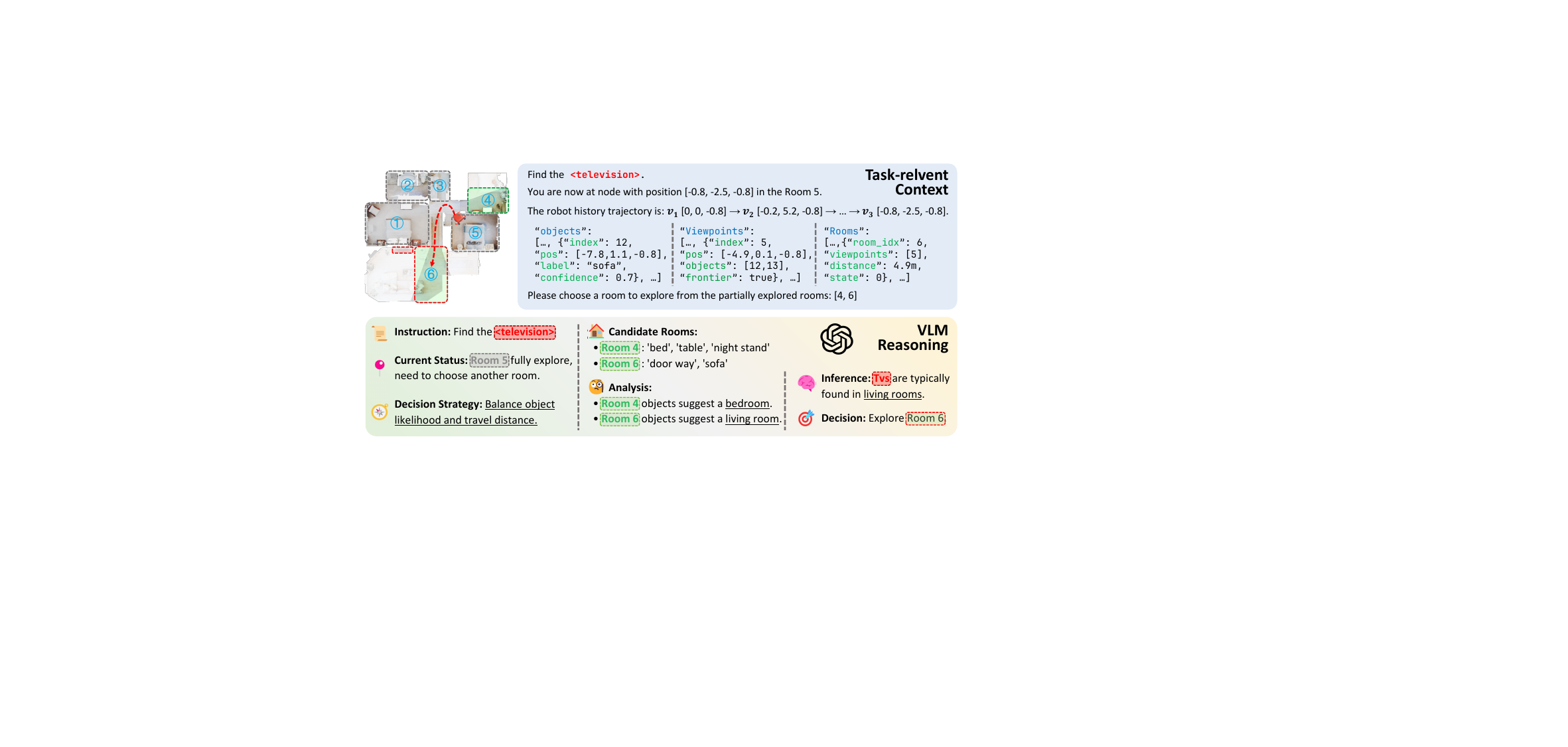}
    \caption{Visualization of the structured prompt and the VLM's reasoning process of selecting the next best room.}
    \vspace{-5pt}
    \label{fig:next_best_room}
\end{figure*}

\textbf{Node:} \zongtai{For efficient exploration, the goal is to control the entire area exhaustively with the minimal number of viewpoint nodes. New viewpoint nodes are preferably selected at locations that maximize the visibility of frontiers.} To select the future viewpoint nodes, we begin by taking the input of agent's position $\mathbf{p}$, coverage range $\zeta_{cover}$, and observed point cloud.
%\zongtai{(project from posed $D$)}
%First, $k$ rays are cast from $\mathbf{p}$ in $k$ uniformly sampled directions to intersect with point cloud and form a polygon $P$ within $\zeta_{cover}$, representing the controlled region of current viewpoint node. Next, we remove the polygon from the point cloud and divide the remaining point cloud into separate regions, which fall into two categories: regions with or without frontiers.
\zongtai{First, we exclude the} \zongtai{region controlled by the current viewpoint node, thereby dividing the remaining point cloud into separate regions, which fall into two categories: regions with or without frontiers.}

\textit{Regions with frontiers:} Inspired by~\cite{yamauchi1997frontier}, we use frontiers to guide viewpoint selection. \zongtai{The detected frontiers are clustered into frontier edge segments, each representing a candidate boundary for exploration. The segments are organized into a graph $G_{frontier}$ (\textcolor{Green}{green dashed lines}), where nodes denote segments and edges denote mutual visibility.}
%The frontiers are identified and clustered into frontier edge segments and construct a graph $G_{frontier}$ (green dashed lines) by connecting every pair of segments that are mutually visible.
%If two segments are visible to each other, an edge is added between them, eventually forming a graph $G_{frontier}$.
% , where nodes represent frontier clusters and edges represent visibility.
Then the \textit{Maximum Clique}
%\footnote{A clique is a subset of vertices in an undirected graph in which every pair of distinct vertices is adjacent, and a maximum clique is one with the largest number of vertices among all such subsets.}
is iteratively removed from $G_{frontier}$. For each removed clique, its center is added as a new viewpoint node $v_i^{vp}$ (\textcolor{Green}{\textit{green node}} in \cref{fig:viewpoint}) to our representation $\mathcal{R}$. \zongtai{This new node serves as the exploration anchor for all frontier segments within the clique.}

\textit{Regions without frontiers:} For these regions, the center of the region is directly added as a new viewpoint node $v_{i}^{vp}$ (\textcolor{Dandelion}{\textit{yellow nodes}} in \cref{fig:viewpoint}) to our representation $\mathcal{R}$.

%\todo{\textbf{Edges between $V^{vp}$:} For each viewpoint node, we check its direct traversability with other viewpoint nodes. If two nodes are direct traversable to each other, an edge is added between them.}
\textbf{Edges between $V^{vp}$:} We evaluate straight-line traversability between each pair of viewpoint nodes. An edge is added if the direct line between two nodes is free of obstacles.
%If so, we add an edge connecting the two nodes.
% 感觉还是有点奇怪

\subsubsection{Object Nodes}
\label{sec:object-node}
% Formally, at each time t, the method detects and segments objects from the RGB-D
% observation and synchronously reconstruct the 3D point clouds for the
% environment. These newly detected objects will be matched with object nodes detected
% from the previous t − 1 frames and merged into them. Objects that do not match
% with any previous one will be registered as a new object node. Each object node has
% a semantic category and its confidence score.
We leverage open-vocabulary detection and segmentation methods~\cite{zhao2024dino,kirillov2023sam} to obtain segmented 3D object instances. Specifically, given the observations at time step $t$, we reconstruct the 3D point cloud of each detected object using the predicted masks, depth map $D_t$ and camera pose $P_t$. For each object, we instantiate an object node at its center, recording attributes such as 3D position, point cloud, predicted label, confidence score and 3D bounding box. Newly instantiated nodes are merged with previously observed nodes if they correspond to the same physical object.

\textbf{Edges between $V^{vp}$ and $V^{obj}$:} An edge is added between $v_i^{vp}$ and $v_j^{obj}$ if $v_j^{obj}$ is within the cover range $\zeta_{cover}$ of $v_i^{vp}$ and is visible from $v_i^{vp}$.
An object can be associated with multiple viewpoints.
If an object isn't connected to any viewpoint, we connect it to the closest visible viewpoint.

\subsubsection{Room Nodes}
\label{sec:room-node}
Following~\cite{werby2024hierarchical, hughes2022hydra}, we identify all walls in the environment and iteratively dilate them to segment the environment into connected components. Then each connected component is added as a room node $v^{room}_i$ to our representation $\mathcal{R}$. Finally, edges are added between each room node and the viewpoint nodes located within the corresponding room.

\begin{algorithm}[H]
    % \fontsize{8}{10}\selectfont  % 设置字号为10pt，行距为12pt
    % \caption*{\fontsize{8}{9}\selectfont \textbf{Alg 1}: Viewpoint Construction}
    \caption*{\textbf{Alg 1}: Viewpoint Construction Process}
    \label{alg:viewpoint-construction}
    \begin{algorithmic}[1]
    \Require Position $p$, Point Cloud $\mathcal{P}$
    \Ensure Updated Viewpoint Nodes $\mathcal{V}^{vp}$
    \State Calculate controlled region $P_c$
    \State $\hat{\mathcal{P}} \gets$ $\mathcal{P} - P_c$
    \State Regions $\mathcal{R}$ $\gets$ \code{Cluster}($\hat{\mathcal{P}}$)
    \For{region $r_i$ in $\mathcal{R}$}
        \If{$r_i$ has no frontiers}
            % \State $v_{i,0}^{vp} \gets \text{center}(r_i)$
            \State $v^{vp} \gets \code{Center}(r_i)$, \; Add new $v^{vp}$ to $\mathcal{V}^{vp}$
        \Else
            \State $\mathcal{C}_i^F \gets \code{FindFrontierCliques}(r_i)$
            % \State Frontier Cliques $\mathcal{C}_i^f$
            % \Statex \hspace{2.5em} $\gets$ Visibility(frontiers)
            \For{$c^F_{i,j}$ in $\mathcal{C}_i^F$}
                \State $v^{vp} \gets \code{Center}(c^F_{i,j})$, \; Add new $v^{vp}$ to $\mathcal{V}^{vp}$
            \EndFor
        \EndIf
    \EndFor
    \end{algorithmic}
\end{algorithm}

\subsection{Object Navigation Policy}
\label{sec:object-navigation-policy}
In this section, we present our efficient two-stage navigation policy, where the VLM performs high-level reasoning and planning at the room level, while the agent conducts fine-grained exploration within each room at the viewpoint level, guided by a VLM-assisted frontier-based strategy and VLM-based target verification.
% for the high-level planning instead of making step-by-step decisions among all reachable viewpoints, the VLM selects the next room to explore based on the spatial layout and semantic information of each room. For low-level exploration within the selected room, we employ a traditional frontier-based algorithm for efficient exploration, while leveraging the VLM to decide whether continued exploration of the current room is worthwhile.
% Additionally, we introduce a early-stop and time-penalized distance mechanism to improve navigation efficiency and VLM Check and Check Again mechanisms to avoid false positives in target detection.
% Additionally, given the importance of accurate target category detection in navigation tasks, we introduce VLM Check and Check Again mechanisms to enhance the reliability of target detection.

\subsubsection{Explore in Room with Early Stop}
\label{sec:explore-in-room}
For efficient low-level exploration within rooms, we introduce VLM-assisted early stop, combining VLM with traditional frontier-based algorithm.
% we first classify frontiers into two types: \textbf{True Frontiers}, which lie along room boundaries indicating incomplete exploration, and \textbf{Inner Frontiers}, which arise from objects' occlusions.
% The agent iteratively navigates to the nearest viewpoint with \textbf{True Frontiers} and explores until all \textbf{True Frontiers} are cleared. If \textbf{Inner Frontiers} still remain in the current room, we query the VLM to decide whether further exploration inside this room is necessary.
We first classify frontiers into two types: \textit{True Frontiers}, which lie along room boundaries indicating incomplete exploration, and \textit{Inner Frontiers}, resulting from objects' occlusions.
The agent iteratively navigates to the nearest viewpoint with \textit{True Frontiers} and explores until all \textit{True Frontiers} are cleared. If \textit{Inner Frontiers} still remain in the current room, we query the VLM to decide whether further exploration inside this room is necessary.
% \textbf{Early Stopping:}
% We classify the frontiers into two categories: \textit{True Frontiers} and \textit{Inner Frontiers}. True frontiers are those that lie on the boundary of the room, indicating the room is not fully explored, while inner frontiers are those that are within the room and caused by occlusions, meaning the room is fully explored while some objects occlude some areas. We will stop exploring the current room if only small inner frontiers are left in this room.

% \input{tables/main_experiment.tex}
\begin{table*}[t]
\centering
% set the column width
\vspace{5pt}
\setlength{\tabcolsep}{5pt}
\resizebox{\textwidth}{!}{
\begin{tabular}{lcccccccccc}
\toprule
\multirow{2}{*}{Method} & \multirow{2}{*}{Open-Set} & \multirow{2}{*}{Zero-Shot}
& \multicolumn{2}{c}{HM3D-v1} & \multicolumn{2}{c}{HM3D-v2}
& \multicolumn{2}{c}{RoboTHOR} & \multicolumn{2}{c}{MP3D} \\
\cmidrule(lr){4-5}\cmidrule(lr){6-7}\cmidrule(lr){8-9}\cmidrule(lr){10-11}
 & & & \textbf{SR (\%)} $\uparrow$ & \textbf{SPL (\%)} $\uparrow$
 & \textbf{SR (\%)} $\uparrow$ & \textbf{SPL (\%)} $\uparrow$
 & \textbf{SR (\%)} $\uparrow$ & \textbf{SPL (\%)} $\uparrow$
 & \textbf{SR (\%)} $\uparrow$ & \textbf{SPL (\%)} $\uparrow$ \\
\midrule
SemEXP~\cite{chaplot2020object}      & \ding{55} & \ding{55}  & -                 &-                  & -                 & -                  & -                 & -                  & 36.0              & 14.4               \\
PONI~\cite{ramakrishnan2022poni}        & \ding{55} & \ding{55}  & -                 &-                  & -                 & -                  & -                 & -                  & 31.8              & 12.1               \\
ZSON~\cite{majumdar2022zson}        & \ding{51} & \ding{55}  & 25.5              &12.6               & -                 & -                  & -                 & -                  & 15.3              & 4.8                \\
L3MVN~\cite{Yu_2023}       & \ding{55} & \ding{51}  & 50.4              &23.1               & 36.3                 & 15.7                  & 41.2              & 22.5               & 34.9              & 14.5               \\
ESC~\cite{zhou2023escexplorationsoftcommonsense}         & \ding{51} & \ding{51}  & 39.2              &22.3               & -                 & -                  & 38.1              & 22.2               & 28.7              & 11.2               \\
VoroNav~\cite{wu2024voronav}     & \ding{51} & \ding{51}  & 42.0              &26.0               & -                 & -                  & -                 & -                  & -                 & -                  \\
VLFM~\cite{yokoyama2024vlfm}        & \ding{51} & \ding{51}  & 52.5&{\ul 30.4}& 63.6& {\ul 32.5}& -                 & -                  & 36.4& {\ul 17.5}\\
VLFM$^\sharp $~\cite{yokoyama2024vlfm}        & \ding{51} & \ding{51}  & 50.9              &23.6               & 56.9                 & 27.5                  & -                 & -                  & 32.5              & 15.9               \\
SG-Nav~\cite{yin2024sg}      & \ding{51} & \ding{51}  & 54.0              &24.9               & 49.6                 & 25.5                  & {\ul 47.5}& {\ul 24.0}& 40.2              & 16.0               \\
OpenFMNav~\cite{kuang2024openfmnav}   & \ding{51} & \ding{51}  & 54.9              &24.4               & -                 & -                  & 44.1              & 23.3               & 37.2              & 15.7               \\
TriHelper~\cite{zhang2024trihelper}   & \ding{51} & \ding{51}  & {\ul 56.5}&25.3               & -                 & -                  & -                 & -                  & -                 & -                  \\
% OSG~\cite{loo2024open}          & \ding{51} & \ding{51} & 69.3              & 28.3               & -                 & -                  & -                 & -                  \\
% CogNav~\cite{cao2024cognav}      & \ding{51} & \ding{51} & 72.5              & 26.2               & 54.6              & 24.3               & 46.6              & 16.1               \\
InstructNav~\cite{longinstructnav} & \ding{51} & \ding{51}  & -                 &-                  & 58.0              & 20.9               & -                 & -                  & -                 & -                  \\
 DORAEMON~\cite{gu2025doraemon}& \ding{51} & \ding{51}  & 55.6& 21.4& {\ul 66.5}& 20.6& -& -& {\ul 41.1}&15.8\\ \midrule
\rowcolor{gray!15} \textbf{\ourmethodplain} & \ding{51} & \ding{51}  & \textbf{62.9}     &\textbf{34.2}      & \textbf{79.6}     & \textbf{38.7}      & \textbf{68.1}     & \textbf{36.3}      & \textbf{52.3}     & \textbf{23.1}      \\
 \rowcolor{gray!15} \textbf{STRIVE$^*$} & \ding{51} & \ding{51}  & \textbf{72.7}     & \textbf{38.2}      & -                 & -                  & \textbf{-}     & \textbf{-}      & \textbf{-}     &\textbf{-}      \\ \bottomrule
\end{tabular}
}
\caption{Comparison with SOTA methods with different settings on HM3D-v1, HM3D-v2, RoboTHOR, and MP3D datasets. We report the Success Rate (SR) and Success weighted by Path Length (SPL) metrics. \icra{Best results are \textbf{in bold}, second best are {\ul underlined}. VLFM$^\sharp$ replaces pre-trained PointNav module with a shortest-path planner for fair comparison. STRIVE$^*$ denotes evaluation restricted to episodes where agent’s starting position and the target object are located on the same floor.}}
\label{tab:main_experiment}
\vspace{-10pt}
\end{table*}
\subsubsection{Next Best Room}
\label{sec:room}
In situations where exploration of the current room is completed \textit{without} finding the target object, we must determine the next room to explore. To guide this decision, we leverage VLM’s commonsense reasoning abilities by providing task-relevant context and general exploration heuristics.
The task-relevant context is consolidated into a combined \textit{Prompt} as described in \cref{fig:next_best_room}, which contains
\textit{1)} target object instruction,
\textit{2)} agent's current state,
\textit{3)} agent's navigation history, and
\textit{4)} the environment representation $\mathcal{R}$ formatted as a JSON file.

Besides the task-relevant context, we also provide the VLM with general exploration heuristics. Specifically, we explicitly instruct the VLM to evaluate two factors: \textit{1)} The semantic similarity between the objects in each room and the target object. \textit{2)} The distance from the agent's current position to each room, aiming to optimize the exploration path by minimizing unnecessary backtracking. Using a Chain-of-Thought reasoning strategy, the VLM selects the most suitable unexplored room for further exploration. Finally, the viewpoint closest to the current position in the selected room is chosen as the next action viewpoint.

Notably, in later navigation stage, continuing forward is more effective than backtracking, as remaining steps may not allow long detours. In light of this, we introduce a penalized distance that weights the geodesic distance by factors reflecting the steps already taken and the number of explored viewpoints along the path to each candidate room.

\subsubsection{VLM-Based Target Verification}
\label{sec:object-found}
% Accurate target category detection is crucial in object navigation. Relying solely on MM-Grounding-DINO has two main drawbacks: (1) the open-vocabulary setting introduces lots of labels with similar meanings, leading to confusion in target detection, and (2) the agent may observe the object from an unfavorable perspective, resulting in incomplete or unclear visual input. We propose two mechanisms to address these issues:
Accurate detection of the target object is crucial in object navigation. However, relying solely on detectin model~\cite{zhao2024dino} often results in false positives. To address this, we propose incorporating the VLM to verify detected target objects, leveraging its ability to reason about the contextual information of the surrounding environment.

% Detect: 当我们检测到了一个目标物体，我们会立刻将该物体与该物体周边环境的RGB image送入VLM中进行验证。VLM会结合该物体和周围环境的语义信息来进一步判断该物体是否为目标物体。

% 走过去时：我们初次检测到一个目标物体时，我们可能没有很好的观测到物体，因为遮挡或者过远的距离，导致该物体的语义信息不准确，这要求我们在更优的视角下，对物体进行观测。与baseline方法从多个视角对物体观测不同，我们在从当前位置走到目标物体的路径上，计算出一个最优的视角位置，在该视角位置上，再次调用VLM对该物体进行验证。这样做的好处是，我们可以在不损失导航效率的前提下，进一步提升目标物体的检测精度。
\textbf{Context-Aware Verification:}
When the agent detects a potential target object,
we prompt the VLM with the detected object and its surrounding visual context for verification.
% we crop the detected region along with its surrounding context and send it to the VLM for verification.
The VLM leverages both the object's appearance and its surrounding semantic information to determine whether it matches the target category, e.g. recognizing a painting of plant as a `decoration' rather than `plant'.

%When the agent detects a potential target object, we send the detected object and its surrounding RGB image to the VLM for verification. The VLM combines the semantic information of the object and its surrounding environment to determine whether the object is indeed the target.

\textbf{Viewpoint-Optimized Re-Verification:}
The agent may initially observe and detect the target object from a suboptimal viewpoint (e.g., under occlusion or from a long distance), resulting in inaccurate detection. To address this, we perform a second observation from a better viewpoint.
%When we first detect a target object, we may not have a good view of it due to occlusion or far distance, leading to inaccurate semantic information.This requires us to observe the object again from a better viewpoint.
Unlike the baseline method~\cite{yin2024sg}, which further observes the target object from multiple viewpoints, we compute the optimal viewpoint along the path from the current position to the target object and only perform one VLM re-verification at that viewpoint. This strategy improves detection accuracy without sacrificing navigation efficiency.

% \textbf{VLM Check:}
% % dino内的部分是不是就可以不要讲了
% We augment the original detection pipeline with a single-category detection process for the target category $C$. First, MM-Grounding-DINO is applied specifically for $C$ to reduce interference from similar categories. For each detected box, we crop the most relevant region from the panaramic image $I$, and feed it to the VLM for further verification. The VLM helps disambiguate context—for example, distinguishing a painting of flowers as a 'decoration' rather than 'flowers'. An object is labeled as category $C$ only when both MM-Grounding-DINO and the VLM agree. Further details are provided in the Appendix.

% \textbf{Check Again:}
% Even after VLM Check, false positives may still occur due to poor viewpoints or long distances. In navigation tasks, since the agent can move freely, we leverage this capability to re-verify the target from a better viewpoint. Specifically, after detecting a potential target, we compute a path toward it and identify a location along the path with optimal view. Upon reaching this location, we gather visual inputs and perform another VLM check. Further details are provided in the Appendix.

% \vspace{-5pt}
\section{Experiment}
\label{sec:experiment}
% \vspace{-5pt}
\begin{figure*}[t]
    \centering
    \vspace{5pt}
    \includegraphics[width=0.98\linewidth]{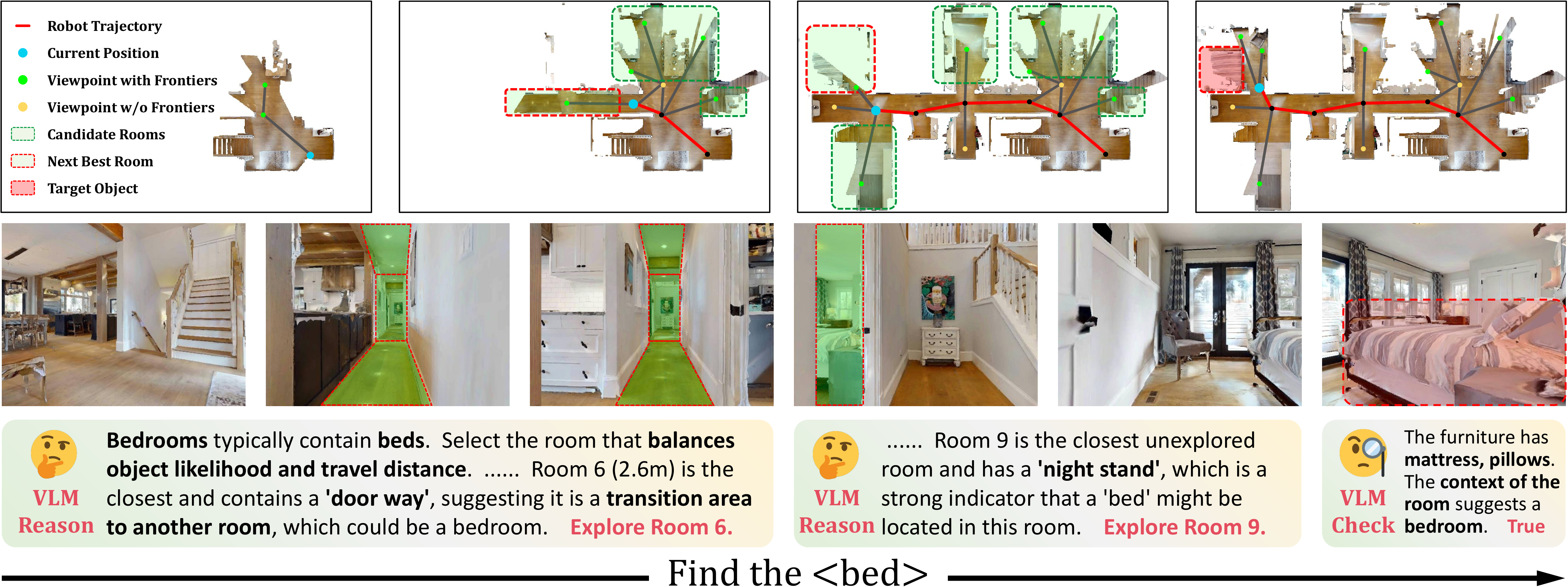}
    \vspace{5pt}
    \caption{\textbf{Qualitative visualization of \ourmethodplain.}
    % In the first two steps, the VLM selects Room 6 and Room 9 based on semantic cues (e.g., `doorway' and `nightstand') and proximity to the agent, balancing object likelihood and travel cost.
    The 1st\&2nd steps show the VLM's reasoning process, where it selects Room 6 and 9 by jointly considering room-layout (`doorway'), semantic cues (`nightstand') and travel cost (penalized distance).
    % In the first two steps, the VLM selects Room 6 and Room 9 balancing object likelihood (semantic cues: e.g., `doorway' and `nightstand') and travel cost(penalized distance to candidtae rooms).
    The final step shows VLM-based verification, using contextual cues (e.g., mattress, pillows) to confirm the target object.}
    \vspace{-10pt}
    \label{fig:qualitative}
\end{figure*}
\subsection{Experiment Setup}
\label{sec:experiment-setup} We evaluate our method by comparing it with state-of-the-art
methods on the Object Navigation task in Habitat~\cite{puig2023habitat3}
simulator. We also conduct real-world experiments across diverse environments.

% \textbf{Dataset:} We perform simulated experiments on 3 datasets: 1) HM3D~\cite{ramakrishnan2021hm3d},
% a large-scale 3D indoor scene dataset comprising 20 high-fidelity scenes with 6 target object categories. 2) 
% MP3D~\cite{Matterport3D}, another 3D scene dataset featuring 11 high-fidelity
% scenes  with 21 target object categories. 3) RoboTHOR~\cite{RoboTHOR}, a 3D indoor scene dataset containing
% 15 scenes with 12 target object categories.

\textbf{Dataset:} We perform simulated experiments on 4 datasets: 1) HM3D-v1~\cite{habitatchallenge2022}, including 2000 episodes across 20 high-fidelity scenes with 6 target object categories. 2) HM3D-v2~\cite{puig2023habitat3}, including 1000 episodes across 36 high-fidelity scenes with 6 target object categories. 3) RoboTHOR~\cite{RoboTHOR}, including 1800 episodes across 15 scenes with 12 target object categories. 4) MP3D~\cite{Matterport3D}, including 2195 episodes across 11 scenes with 21 target object categories.

\textbf{Evaluation Metrics:} Following \cite{habitatchallenge2023}, we use 4 metrics
to evaluate the performance: 1) Success Rate (SR): the percentage of episodes in
which the agent reaches the target object within a success distance. 2) Success
weighted by Path Length (SPL): the success rate weighted by the ratio of the shortest
path length to the actual path length. 3) Distance to Goal (DTG): the final
distance to the target object at the end of the episode. 4) SoftSPL: replacing the binary success term in SPL with a “soft” value that indicates the progress made by the agent towards the goal.

\textbf{Implementation Details:}
Each episode allows a maximum of 500 steps and a success distance of 1.0m. The agent observes the environment using a $640 \times 480$ RGB-D image, with depth values from 0.5m to 5.0m and a horizontal field of view (HFOV) of $79^\circ$. The camera is mounted 0.88m above the ground. The agent moves forward by 0.25m per step and rotates by $30^\circ$.
For VLM, we use Gemini~\cite{team2023gemini} (gemini-2.0-flash), and for object detection and segmentation, we use MM-GroundingDINO~\cite{zhao2024dino} and SAM~\cite{kirillov2023sam}.
All experiments are conducted on RTX4090 GPUs.
% Please note that our method is a purely zero-shot method and can generalize to any unseen environments and objects without additional training.

For real-world experiments, we deploy \ourmethod on the Mecanum wheel platform~\cite{autonomy_stack_mecanum_wheel}, which is equipped with a Ricoh Theta Z1 360-degree camera for RGB image capturing and a Livox Mid-360 LiDAR for 3D point cloud acquisition. To maintain compatibility with the input format in simulation, the collected point clouds are converted into depth maps when necessary.

\subsection{Quantitative Results in Simulator}
% \vspace{-5pt}
\label{sec:quantitative-results}
We compare \ourmethod with state-of-the-art object navigation methods in different settings
%, including open-set,closed-set and supervised, zero-shot methods
as shown in \cref{tab:main_experiment}. \ourmethod significantly outperforms all baselines across all benchmarks. \icra{Compared to the second-best methods, it achieves improvements of +6.4\% SR / +3.6\% SPL on HM3D-v1, +13.1\% SR / +6.2\% SPL on HM3D-v2, +20.6\% SR / +12.3\% SPL on RoboTHOR, and +11.2\% SR / +5.5\% SPL on MP3D.}
\zongtai{For HM3D-v1, since our method is designed for single-floor object navigation, we additionally evaluate only on episodes where the agent’s starting position and the target object are on the same floor. Under this setting, \ourmethod achieves 72.7\% SR and 38.2\% SPL.} % 我觉得不需要特地写这一段，我们的方法也不是只是为了单层设计的，只是没仔细实现。理论上把z坐标做好是可以支持多层的。
Overall, %the improvement in SPL is more significant than that in SR, indicating
\zongtai{the improvement in SPL shows} that our representation and navigation policy effectively improve navigation efficiency.
Furthermore, increased navigation efficiency enables the agent to explore a larger area within a limited number of steps. The improvement in SR results from the combined effects of more efficient navigation, better utilization of the VLM's reasoning capabilities, and more accurate VLM-based verification.
% \zongtai{这样说有点歧义，好像是单纯因为提高了效率探索更多才提高SR}
% \vspace{-5pt}
\subsection{Qualitative Results in Simulator}
% \vspace{-5pt}
\label{sec:qualitative-results}
We visualize the navigation process of \ourmethod in \cref{fig:qualitative}. The results demonstrate that our structured representation enables the VLM to reason effectively about both spatial layout (e.g., a room with a "doorway" can lead to other rooms) and semantic cues (e.g., nightstands suggesting bedrooms), leading to improved room selection. Furthermore, the VLM balances the likelihood of finding the target object against travel distance cost when planning room-to-room exploration. It also leverages contextual information to re-verify detected objects and effectively reduces false positives.

\begin{table}[t]
    \centering
    \resizebox{0.48\textwidth}{!}{
    \begin{tabular}{c|ccccc|c}
    \toprule
    Object Category         & Sofa  & Chair & Monitor & Table & Garbage bin & Average \\ \midrule
    Success Rate(\%) & 85    & 90    & 70      & 85    & 75          & 81      \\
    Runtime(s)         & 68.26 & 45.22 & 70.86   & 68.80 & 62.53       & 63.13  \\ \bottomrule
    \end{tabular}
    }
    \caption{\textbf{Real-World Quantitative Results.} We conducted 20 episodes for each object category and report the success rate and average runtime.}
    \vspace{-10pt}
    \label{tab:real_world}
\end{table}

\subsection{Real-World Experiments}
\label{sec:real-world-experiments}
% We deploy our method on the Mecanum wheel platform and conduct 15 real-world experiments in over 10 different environments, including offices, meeting rooms, classrooms, lounges, dining areas, corridors, and kitchens. Examples of experimental scenes and outcomes are shown in Fig.~\ref{fig:enter-label}.
We conduct 120 real-world experiments across 15 diffrent target categories and 10 different environments, including offices, meeting rooms, classrooms, lounges, dining areas, corridors, and kitchens. Part of the experiment environments and results are shown in \cref{fig:teaser}.
Compared to simulation, real-world deployment presents additional challenges. Lidar-captured point clouds are much sparser than depth maps, and real environments are often more cluttered, introducing noise that affects both exploration and object detection. Despite these challenges, our agent demonstrates robust performance.

We elaborate on 2 difficult environments in \cref{fig:teaser}. In the left scenario, the agent is initialized inside a small enclosed room connected to a larger lounge area.
Despite the presence of inner frontiers—regions occluded by furniture—the agent correctly decides to abort exhaustive exploration of this room. Instead, it exits the room early and shifts its attention to unexplored rooms nearby, where it ultimately locates the target object. In the right scenario, the agent is initialized in a dining area and instructed to find a `Garbage bin'. \ourmethod successfully uses the VLM to reason on semantic associations (`refrigerator' and `bins') and find the target object efficiently.

\zongtai{We also evaluated the runtime performance of our method in real-world scenarios. In \cref{tab:real_world}, we present the success rates and average runtimes for 100 episodes over 5 object categories, with 20 episodes for each object category. Our method achieves high success rates and short average times over all 5 categories, demonstrating its effectiveness in the real-world environment.}

% We also showcase selected successful examples from other environments in the figures.
%For other environments, we also present final frames where the agent successfully reaches the target object.
%From these results, we can conclude that \ourmethod can handle diverse and complex real-world environments.
% \vspace{-5pt}
\subsection{Ablation Study}
% \vspace{-5pt}
\label{sec:ablation-study}

\begin{table}[t]
    \centering
    \vspace{5pt}
    \small
    \setlength{\tabcolsep}{2pt}
    \centering
    \resizebox{1.0\linewidth}{!}{%
    \begin{tabular}{ccc|cccc}
    \toprule
    $V^{vp}$ & $V^{obj}$ & $V^{room}$ & SR $\uparrow$ & SPL $\uparrow$ & S-SPL $\uparrow$ & DTG(m) $\downarrow$ \\
    \midrule
    \cmark   & \xmark    & \xmark     & 71.3              & 33.2               & 35.2                 & 1.86                \\
    \cmark   & \cmark    & \xmark     & 72.4              & 34.0               & 36.1                 & 1.95                \\
    \cmark   & \xmark    & \cmark     & 72.9              & 33.8               & 35.4                 & 1.86                \\
    \cmark   & \cmark    & \cmark     & \textbf{75.0}              & \textbf{34.9}               & \textbf{36.5}                 & \textbf{1.80}                \\
    \bottomrule
    \end{tabular}
    }
    \caption{\textbf{Ablation study of representation on HM3D.}
    We adopt a \textit{viewpoint-level navigation policy} for experiment consistency.}
    \vspace{-8pt}
    \label{tab:ablation_rep}
\end{table}
% 请注意在这个实验中，由于实验设置限制，我们使用了一个原始的导航策略，我们让VLM以视点为粒度引导navigation，instead of让VLM以房间为粒度引导navigation。

\textbf{Multi-layer Representation:} We conduct an ablation study to evaluate the contributions of object nodes and room nodes in our multi-layer scene representation. Since room-level navigation depends on room nodes, it cannot be used when they are removed. To ensure consistency, we adopt a basic navigation policy that allows the VLM to guide navigation at the viewpoint level instead of the room level. As shown in \cref{tab:ablation_rep}, both object nodes and room nodes contribute significantly to performance improvement.
% \todo{below}
%\todo{这里的实验没法说明下面的这些定性的事情，最好在填完了表之后根据表里的结果来写。或者不写。}
Incorporating object nodes improves the agent's ability to localize target objects, while adding room nodes enhances its understanding of the environment layout. Using both layers together leads to a substantial improvement over using either individually.
%Specifically, the object node enhances the agent's ability to identify target objects, while the room node aids in understanding the environment layout. The combination of both nodes leads to a significant performance boost.

\textbf{Navigation Policy:}
We conduct ablation studies to evaluate the effectiveness of our navigation policy, VLM-assisted early stopping, penalized distance, and VLM-based verification, as summarized in \cref{tab:ablation_policy}. The last two rows compare our room-level planning policy against a basic viewpoint-level approach. Results show that room-level planning enables the agent to better leverage the VLM's reasoning capabilities, significantly boosting performance. We also report the average VLM token usage per episode. By querying the VLM only for room-level planning, our method significantly reduces token consumption compared to viewpoint-level planning. Finally, the VLM-assisted early stopping, penalized distance, and verification modules each contribute to further performance gains.

% \input{tables/ablation_all.tex}

% In 12 relatively simple scenarios, where the target is within 5 meters of the agent's initial position, our agent achieves a 100\% success rate. We also test the agent in 3 more challenging settings, where the initial distance to the target ranges from 10 to 15 meters, and the target is not visible from the starting position. The agent must explore and reason based on scene context to find the target. In these cases, we conduct multiple repeated trials and observe a success rate exceeding 80\%, demonstrating the agent’s ability to explore, reason over the scene context, and efficiently locate the target, as illustrated in Fig.~\ref{fig:enter-label}.

\section{Conclusion}
% \vspace{-5pt}
\label{sec:conclusion}
In this paper, we introduce \ourmethod, a novel framework that incrementally constructs a structured scene representation and leverages VLM's reasoning capabilities to achieve efficient object navigation. \ourmethod incrementally builds a multi-layer representation of the environment, consisting of room, viewpoint and object nodes. Based on this representation, we design an efficient two-stage VLM-guided navigation policy, which leverages VLM reasoning for room-level planning while using VLM together with traditional frontier-based methods for efficient exploration within rooms. To further improve robustness, we incorporate VLM-based target verification, utilizing VLMs' contextual understanding to improve detection accuracy. Experiments across three simulated benchmarks demonstrate that \ourmethod achieves state-of-the-art performance, significantly improving both success rate and navigation efficiency.
% \todo{These results underscore the benefits of structured representation and strategic VLM guidance for scalable, zero-shot object navigation.}
Furthermore, our real-world experiments demonstrate the robustness and practicality of \ourmethod in navigating complex and diverse real-world environments.

\begin{table}[t]
    \centering
    \vspace{5pt}
    \small
    \setlength{\tabcolsep}{3pt}
    \centering
    \small
    \resizebox{1.0\linewidth}{!}{%
    \begin{tabular}{c|ccccc}
    \toprule
    & SR $\uparrow$ & SPL $\uparrow$ & S-SPL $\uparrow$ & DTG(m) $\downarrow$  & Tokens $\downarrow$\\ \midrule
    w/o Early Stop   & 74.8 & 34.8 & 36.4 & 1.62 & - \\
    w/o Penalized Dist   & 73.7 & 36.1 & 36.9 & 1.47 & - \\
    w/o VLM-Verify       & 72.1 & 32.7 & 34.1 & 1.83 & - \\ \midrule
    \rowcolor{gray!15} Viewpoint Policy   & 75.0 & 34.9 & 36.5 & 1.80 & 22935 \\
    \rowcolor{gray!15} \textbf{\ourmethodplain}   & \textbf{79.6} & \textbf{38.7} & \textbf{38.9} & \textbf{1.29} & \textbf{8068} \\
    \bottomrule
    \end{tabular}
    }
    \caption{\textbf{Ablation study of navigation policy components on HM3D.} Viewpoint Policy stands for VLM planning on viewpoint-level.}
    \vspace{-7pt}
    \label{tab:ablation_policy}
\end{table}

%%%%%%%%%%%%%%%%%%%%%%%%%%%%%%%%%%%%%%%%%%%%%%%%%%%%%%%%%%%%%%%%%%%%%%%%%%%%%%%%

%%%%%%%%%%%%%%%%%%%%%%%%%%%%%%%%%%%%%%%%%%%%%%%%%%%%%%%%%%%%%%%%%%%%%%%%%%%%%%%%



% \section*{ACKNOWLEDGMENT}

%%%%%%%%%%%%%%%%%%%%%%%%%%%%%%%%%%%%%%%%%%%%%%%%%%%%%%%%%%%%%%%%%%%%%%%%%%%%%%%%

\bibliographystyle{IEEEtran}
\bibliography{main} % .bib

\clearpage
\appendices
\crefname{section}{App.}{Apps.}
% \appendix
% \renewcommand{\thesection}{\Alph{section}}  % 输出 A, B, C, D 等样式
% \setcounter{section}{0}
\twocolumn[{%
\begin{center}
  {\LARGE\textbf{\ourmethodplain: Structured Representation Integrating\\ VLM Reasoning for Efficient Object Navigation\\[1.1ex](Appendix)}}
\end{center}%
}]

\section{Overview}
    \label{sec:overview}
    In this supplementary material, more details about the proposed \ourmethod and more experimental results are provided, including:
    \begin{itemize}
        \item \textbf{General Exploration Heuristic:} The general exploration heuristic prompt we provide to the VLM to help it make better decisions (\cref{sec:heuristic}).
        \item \textbf{Detailed Structure of Task-relevant Context:} The detailed structure of the Task-relevant Context generated from our proposed representation $\mathcal{R}$ to prompt the VLM with whole environment information for better reasoning (\cref{sec:json_structure}).
        \item \textbf{Detailed Experiment Results:} We provide more detailed experimental results, including the performance of \ourmethod on different categories of objects on HM3D~\cite{ramakrishnan2021hm3d} and more qualitative results (\cref{sec:more_exp}).
        \item \textbf{Examples of VLM Reasoning:} We provide more examples of VLM reasoning results, including the reasoning process and the final decision (\cref{sec:vlm_reasoning}).
        \item \textbf{Details of VLM-based Verification:} We provide a detailed explaination of the VLM-based verification process, including Context-Aware Verification and Viewpoint-Optimized Re-Verification, and we also provide more examples of VLM-based verification results, including the reasoning process and the final decision (\cref{sec:vlm_verification}).
        % 我们简要展示了数据集中的一些错误标注的示例。
        \item \textbf{Dataset Error:} We briefly show some examples of mislabelled data in the dataset (\cref{sec:dataset_error}).
        \item \textbf{Detailed Room Segmentation:} We provide more details about the room segmentation process (\cref{sec:room_seg}).
    \end{itemize}

    \section{Exploration Heuristics}
\label{sec:heuristic}
% 为了让VLM更好的引导navigation以尽快完成当前的任务，我们给VLM提供了一个General prompt来让他对这个任务有整体概念，并向其解释了我们所提供的JSON file里面每个部分的含义，同时explicitly要求VLM在每一次做决定时，都要综合考虑每个candidate 房间内，目标物体出现的概率和选择每个房间进行探索所需要花费的travel cost。
In order to make the VLM better guide the navigation to complete the current task as soon as possible, we provide a general prompt to give it an overall concept of the object navigation task. 
We also explain the meaning of each part in the provided JSON file and explicitly require the VLM to consider the probability of the target object appearing in each candidate room and the travel cost required to explore each room when making decisions.
% 具体来说，我们的prompt包含了以下几个部分：

\begin{lstlisting}[caption={General Object Navigation Heuristic Prompt}]
PROMPT = """
    You are a wheeled mobile robot operating in an indoor environment. Your goal is to efficiently find a target object based on a human-provided instruction in a new house. The current room you are in has been fully explored. To achieve the goal, you must select the next room to explore from the partially explored rooms listed in a JSON file, aiming to complete the task as quickly as possible.

### Provided Information:
    1. A specific instruction describing the task.
    2. A description of your current position and previous trajectories.
    3. A JSON file containing details about the scene, including rooms, viewpoints, and objects.

The JSON file contains the following information:
- **Objects**
    - `object_idx`: A unique identifier for the object.
    - `position`: The spatial coordinates of the object.
    - `class`: The category or type of the object.
    - `confidence`: The confidence level of the classification result.
    - `size`: The bounding box size of the object (in meters).

- **Viewpoints**:
    - `viewpoint_idx`: A unique identifier for the viewpoint.
    - `position`: The spatial coordinates of the viewpoint.
    - `state`: The state of the viewpoint (`1` for visited, `0` for unvisited).
    - `neighbors`: A list of connected viewpoints.
    - `has_frontier`: Relevant only when the viewpoint is unvisited.
    - `True`: The viewpoint has a frontier, meaning unknown regions exist around it.
    - `False`: The area around the viewpoint has already been observed from distant viewpoints, but small objects may still be unclear.
    - `objects`: A list of objects observable from the viewpoint.

- **Rooms**
    - `room_idx`: A unique identifier for the room.
    - `state`: The state of the room (`1` for fully explored, `0` for partially explored).
    - 'distance': The distance (in meters) the robot needs to travel to reach this room.
    - `viewpoints`: A list of viewpoints in the room.

### Task:
    You must carefully analyze the JSON file, using logical reasoning and common sense, to select the next room to explore from the list of partially explored rooms. Consider the following factors:
    - Evaluate how closely each room's viewpoints aligns with the overall task objective.
    - Optimize the exploration path by leveraging the robot's current momentum and minimizing unnecessary backtracking or redundant movements.
    - Assess the likelihood that exploring the selected room will meaningfully advance or complete the overall task.

### Output Format:
    Your response should include:
    - 'steps': The chain of thought leading to the decision.
    - `final_answer`: The `idx` of the next room to explore.
    - `reason`: The rationale for selecting this room.

**Note:** The chosen room must be partially explored.
"""
\end{lstlisting}
    \section{Json Structure}
\label{sec:json_structure}
% 在这里我们提供了用于提示VLM的当前navigation进程和目前已知的环境信息的任务相关上下文的详细描述。
Here we provide a detailed description of the task-related context used to prompt the VLM about the current navigation process and the currently known environmental information.

\begin{figure}[h]
    \centering
    \includegraphics[width=0.95\linewidth]{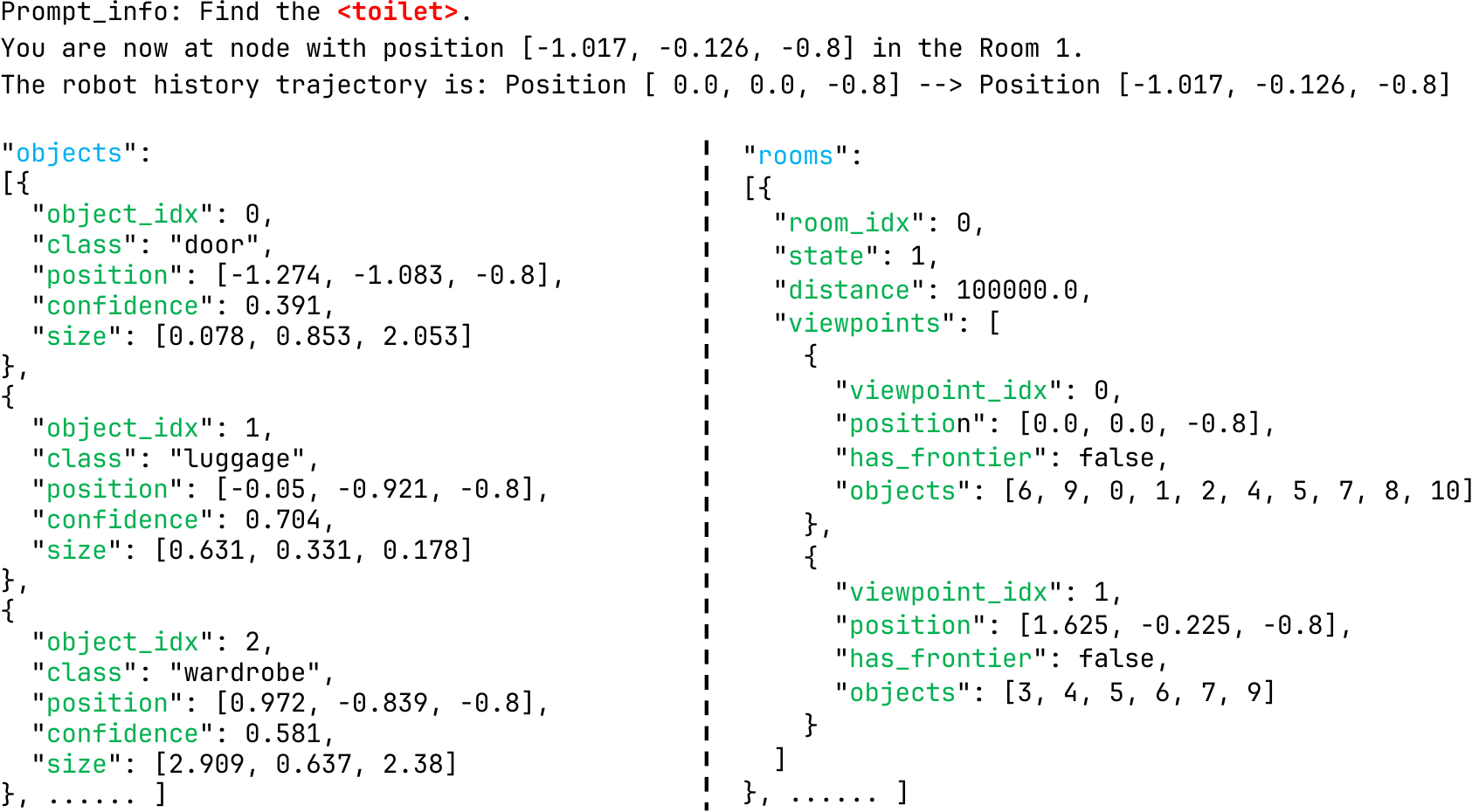}
    \caption{\textbf{Visualization of the Json file.}}
    \label{fig:supp_json}
\end{figure}

% 该上下文包含了以下几个部分：
As shown in \cref{fig:supp_json}, it consists of the following parts:
\begin{itemize}
    \item \textbf{Target Object:} It begins by specifying the target object as \textbf{\texttt{"Find the \textless target object\textgreater"}}.
    \item \textbf{Current Viewpoint and Position:} It then states the robot's current viewpoint and position as \textbf{\texttt{"The robot is now at Viewpoint with position [x,y,z] in Room $r_i$"}}.
    \item \textbf{Navigation History:} The navigation history up to the current step is provided as {\texttt{"The robot history trajectory is Position [x,y,z] $\rightarrow$ , Position [x,y,z] $\rightarrow$ \dots"}}.
    \item \textbf{Scene Representation:} The scene representation $\mathcal{R}$ is formatted as a JSON file as the last part. This representation contains information about the layout and semantic information of the environment, which is crucial for the VLM to make informed decisions. The JSON file is structured as the format of Rooms-Viewpoints-Objects. For detailed json structure, please refer to \cref{fig:supp_json}.
\end{itemize}

Please note that when translating our representation $\mathcal{R}$ into a JSON file, we begin by listing all the objects in the scene. This is because an object may be associated with multiple viewpoints; directly listing objects under each viewpoint would lead to redundancy and may exceed the prompt's length limit.
    \section{More Experiment Results}
\label{sec:more_exp}
We provide more experimental results on the HM3D dataset. First we show the Success Rate on each target object category of the HM3D dataset in Table \ref{tab:hm3d_category}. We can see that our method achieves the best performance on most of the categories. 

\begin{table}[h]
    \centering
    \resizebox{0.99\linewidth}{!}{%
    \begin{tabular}{l|cccccc|c}
    \toprule
    Method                     & bed  & chair & plant & sofa & toilet & tv\_monitor & Average \\ \midrule
    L3MVN~\cite{Yu_2023}         & 52.9 & 51.6  & 46.4  & 50.1 & 41.5   & 54.2        & 49.5    \\
    TriHelper~\cite{zhang2024trihelper} & 57.1 & 58.6  & 58.3  & 58.9 & 52.3   & 57.4        & 57.1    \\
    % CogNav~\cite{cao2024cognav}               & {\ul 67.9}    & {\ul 73.4}     & \textbf{73.1} & {\ul 67.0}    & {\ul 72.6}    & \textbf{74.0} & {\ul 72.5}    \\
    \ourmethodplain (ours) & \textbf{83.8} & \textbf{86.2} & {\ul 67.6}    & \textbf{81.2} & \textbf{81.9} & {\ul 73.3}    & \textbf{79.6} \\
    \bottomrule
    \end{tabular}}
        \caption{\textbf{Success rate of each category on HM3D~\cite{ramakrishnan2021hm3d}.} The best and second best results are highlighted in \textbf{bold} and \underline{underline}, respectively.}
    \label{tab:hm3d_category}
\end{table}

\begin{figure*}[t]
    \centering
    \includegraphics[width=0.77\linewidth]{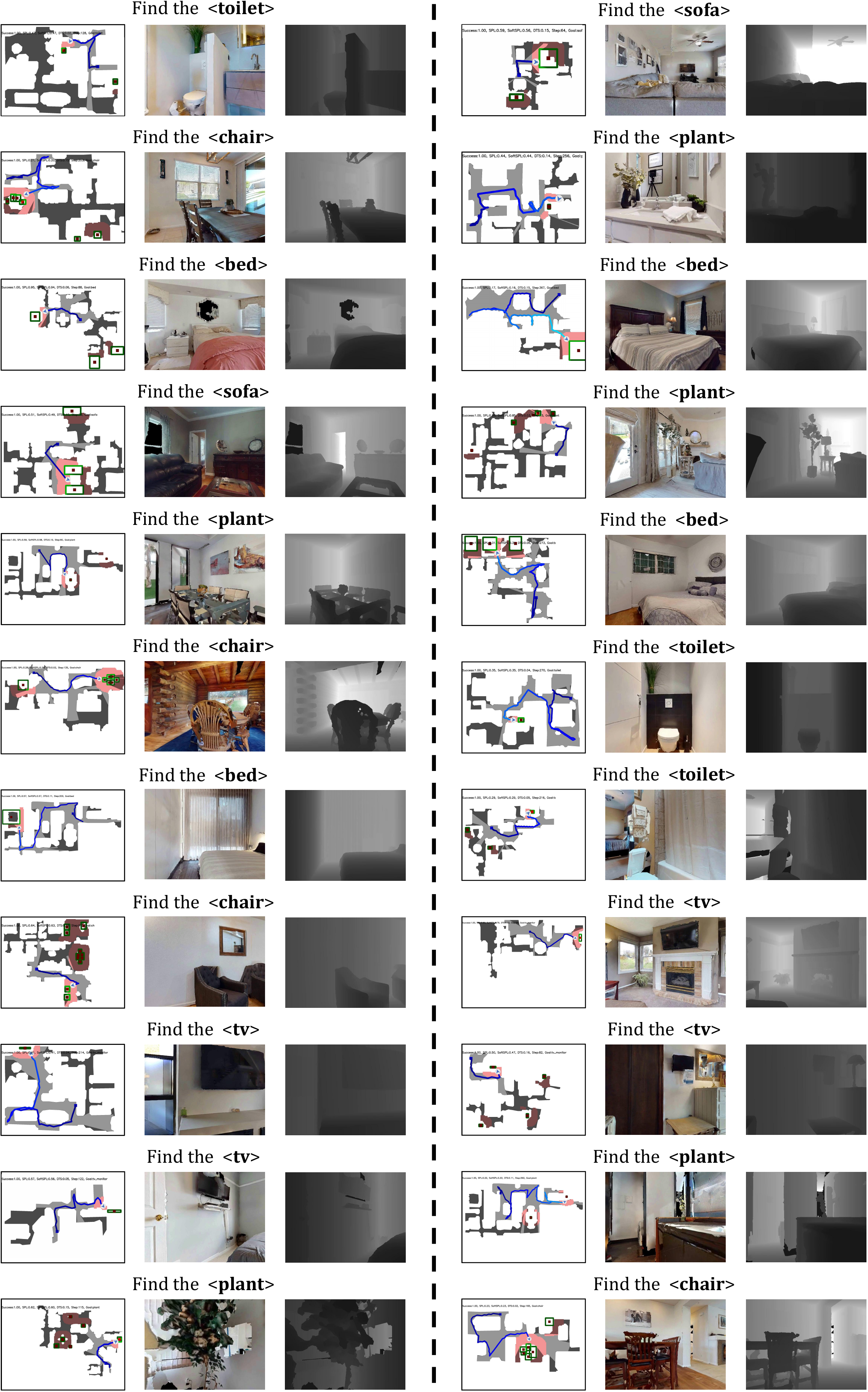}
    \caption{\textbf{Qualitative results of \ourmethodplain on HM3D.} We show the trajectory of the agent and the final RGB-D image when the agent stops. }
    \label{fig:more_qualitative}
\end{figure*}

We also show more qualitative results of our method on the HM3D dataset in Figure \ref{fig:more_qualitative}. We visualize the trajectory of the agent and the final RGB-D image when the agent stops. The agent is able to efficiently navigate to the target object and stop at a reasonable viewpoint to observe the object. 

% 我们展示了agent的轨迹以及最终agent停止时的RGB-D图像。

\clearpage
    \clearpage
\section{Examples of VLM Reasoning}
\label{sec:vlm_reasoning}
\begin{figure*}[b]
    \centering
    \includegraphics[width=0.77\linewidth]{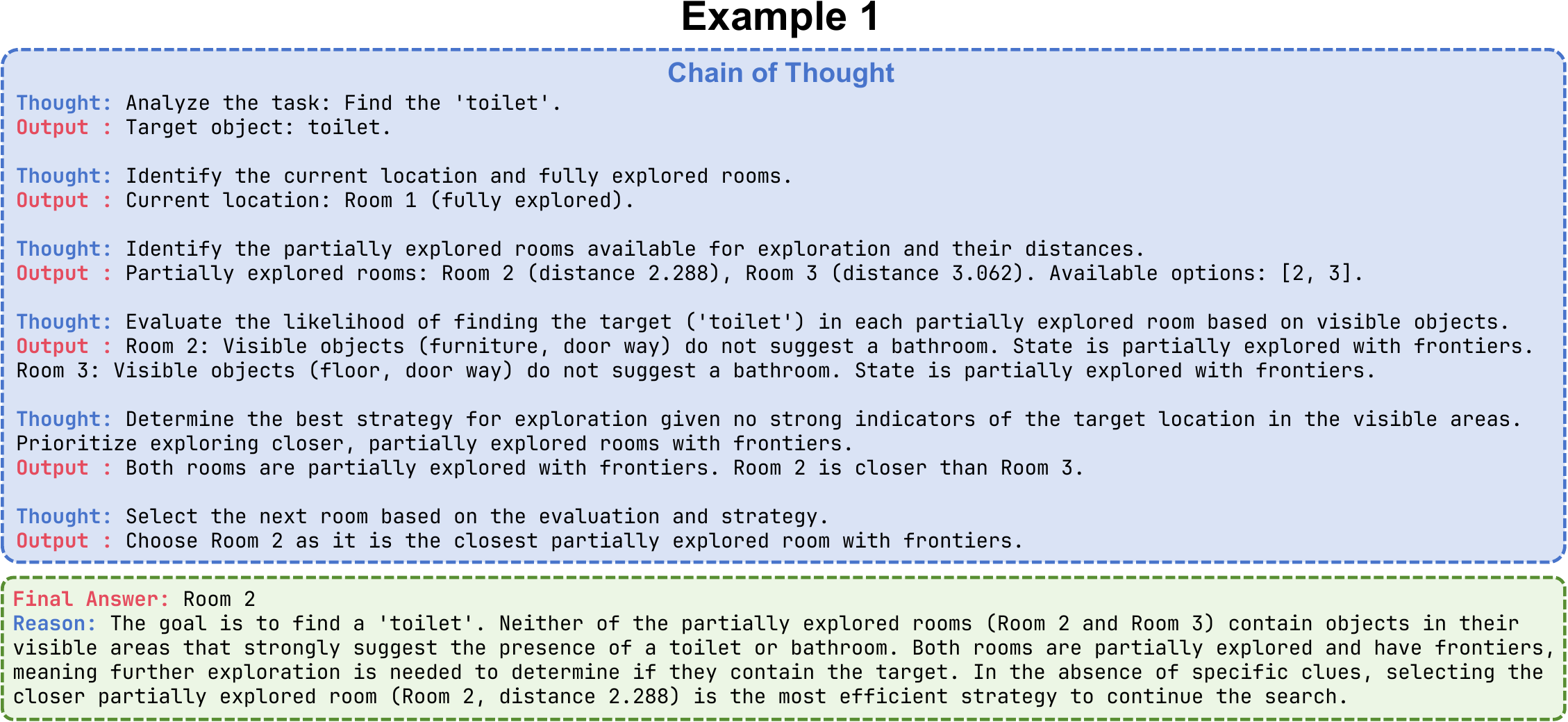}
    \caption{\textbf{Example 1 of VLM reasoning.} }
    \label{fig:reason1}
\end{figure*}

\begin{figure*}[b]
    \centering
    \includegraphics[width=0.77\linewidth]{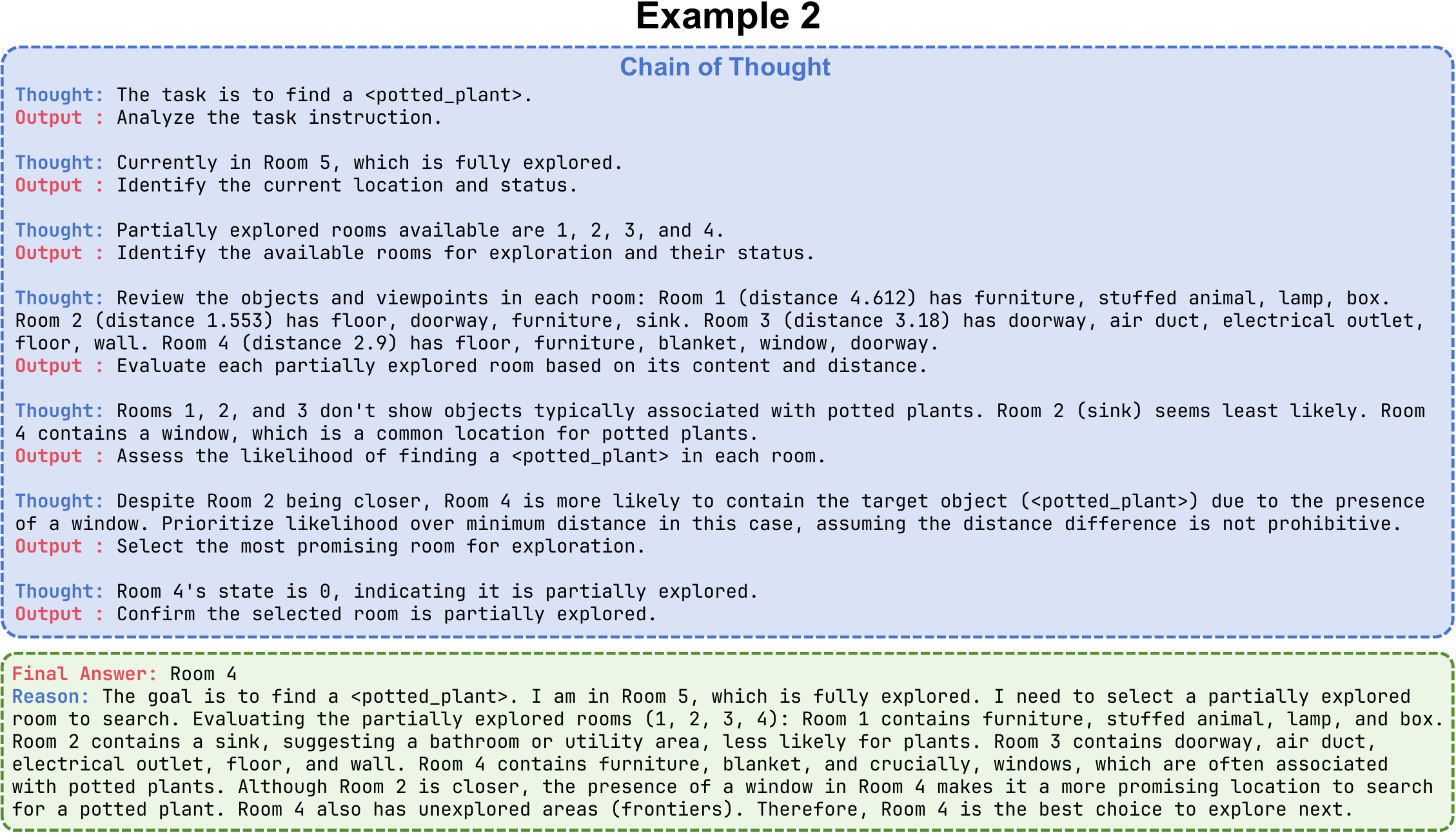}
    \caption{\textbf{Example 2 of VLM reasoning.} }
    \label{fig:reason2}
\end{figure*}

\begin{figure*}[b]
    \centering
    \includegraphics[width=0.77\linewidth]{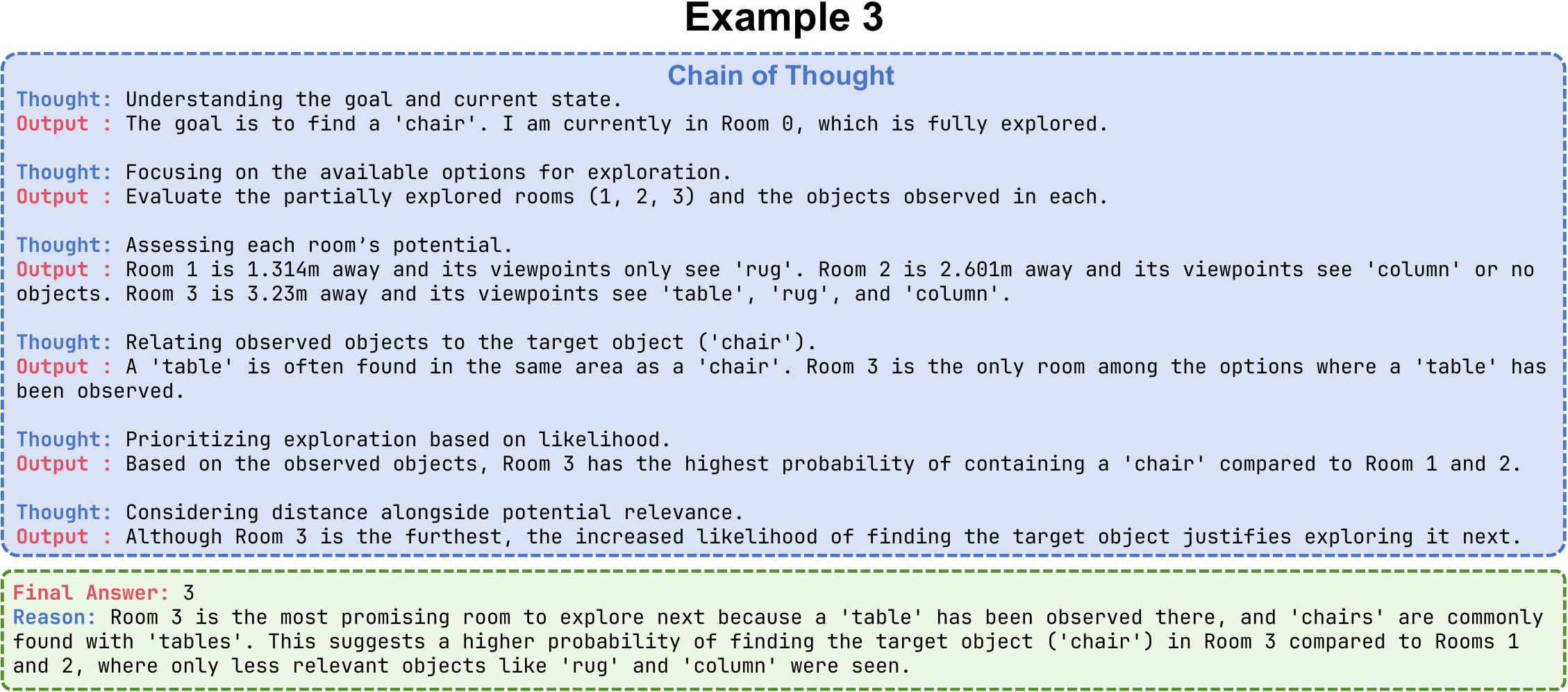}
    \caption{\textbf{Example 3 of VLM reasoning.} }
    \label{fig:reason3}
\end{figure*}

We provide additional examples of the VLM’s reasoning process in \cref{fig:reason1,fig:reason2,fig:reason3}. These results demonstrate that our structured representation enables the VLM to reason effectively over both spatial layout and semantic cues, leading to more accurate room selection. Moreover, the VLM is able to balance the likelihood of finding the target object with the travel distance cost when planning room-to-room exploration.
% \begin{figure}[ht]
%     \centering
%     \includegraphics[width=0.99\linewidth]{figures/vlm_reason.pdf}
%     \caption{\textbf{Examples of VLM reasoning.} }
%     \label{fig:reason}
% \end{figure}

% \clearpage
    \section{Details of VLM-based Verification}
\label{sec:vlm_verification}
\subsection{Context-Aware Verification}
\begin{figure*}
    \centering
    \includegraphics[width=0.85\linewidth]{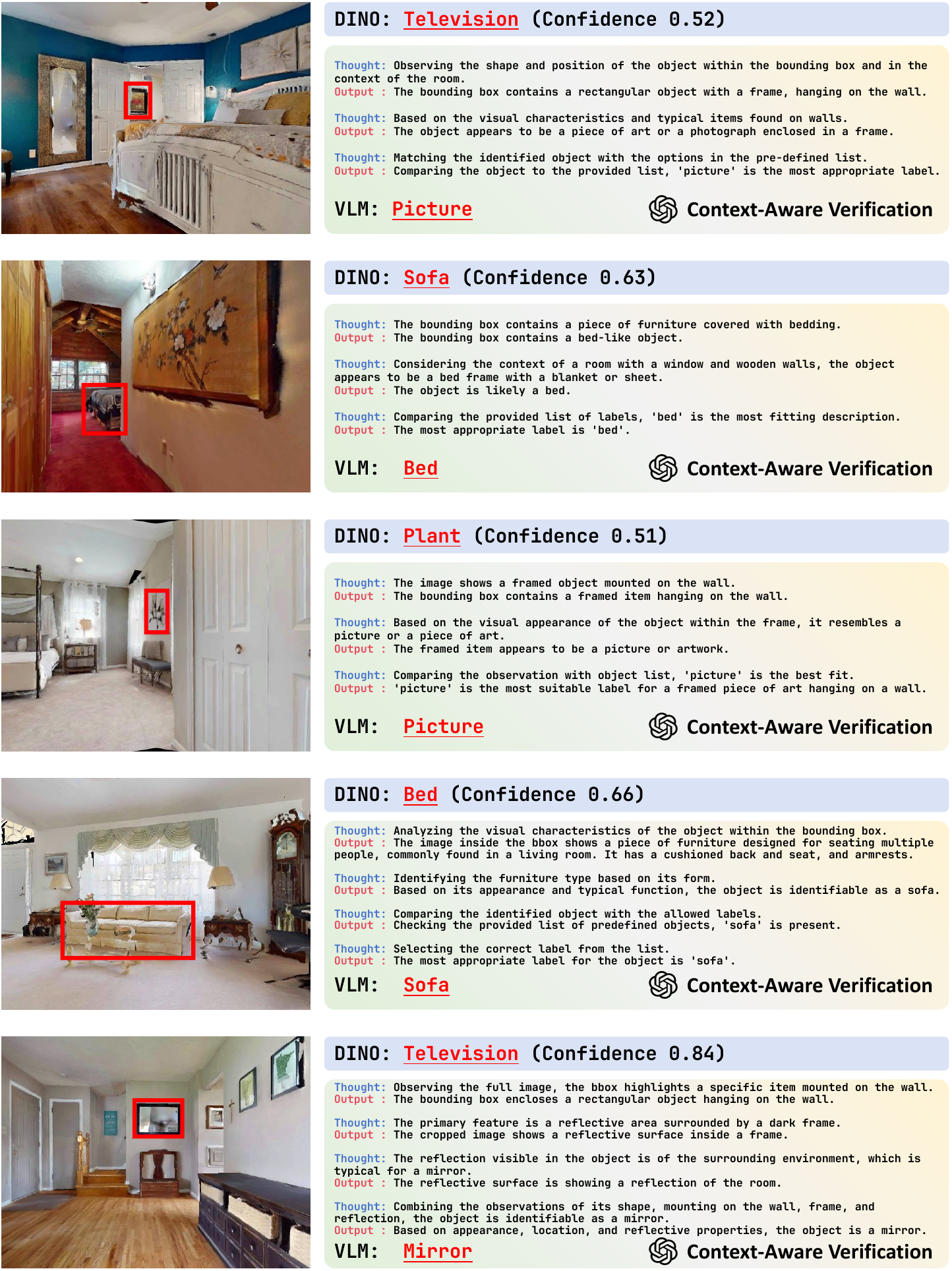}
    \caption{\textbf{Examples of context-aware verification.}}
    \label{fig:context_aware_verification_example}
\end{figure*}

For the context-aware verification, we provide more examples in \cref{fig:context_aware_verification_example}. These examples show that the VLM can effectively utilize the surrounding context information to verify the detected target objects and avoid false positives. 

\subsection{Viewpoint-Optimized Re-Verification}
For the viewpoint-optimized re-verification, we first introduce how to compute a better viewpoint for the re-verification. 
% 我们一旦检测到了一个目标物体，我们会规划一条从当前位置到目标物体的路径，并在这个路径上以agent的前进步长（0.25m）为间距采样一系列的点作为候选的视角点。对于每个时间点，我们计算在该视角点下的目标物体在相机画幅中可见部分的bounding box的w，h，以及目标物体可见部分的fraction。我们倒序遍历这条路径上的每个采样点（从物体到现在位置），我们选择满足条件的第一个点作为re-verification的视角点。我们选择的条件是：1）在该视角点下，能看到目标物体fraction>95%；2）该视角点下的w > \alpha * w_{ori}，h > \alpha * h_{ori}，其中\alpha=1.5，w_{ori}和h_{ori}分别是原始视角下的目标物体的bounding box的w和h。我们选择这个条件是因为我们希望在re-verification的视角下，既要看到大部分都目标物体，同时也要保证对于目标物体有一个足够的的视角范围，这样才能保证VLM能够更好地进行re-verification。
Once we detect a target object, we plan a path from the current position to the target object and sample a series of points along this path with a step size of 0.25m (the agent's forward step length) as candidate viewpoints. For each sampled point, we compute the width (w), height (h), and visible fraction of the target object in the camera frame when the agent faces directly towards the target object. We traverse the sampled points in reverse order (from the target object back to the current position) and select the first point that satisfies all the following conditions as the re-verification viewpoint:
\begin{enumerate}
    \item More than 95\% of the target object's point cloud is visible from this viewpoint.
    \item The 2D bounding box area of the target object exceeds that at the original viewpoint.
    \item The estimated object width and height at this viewpoint are both greater than 80\% of their original values, scaled by the square root of the bounding box area ratio.
\end{enumerate}
The square root of the area ratio is adopted as a scaling factor to account for the expected increase in apparent object dimensions under improved viewing conditions, thereby ensuring a broader and clearer observation for more reliable VLM-based re-verification.

\begin{figure*}
    \centering
    \includegraphics[width=0.95\linewidth]{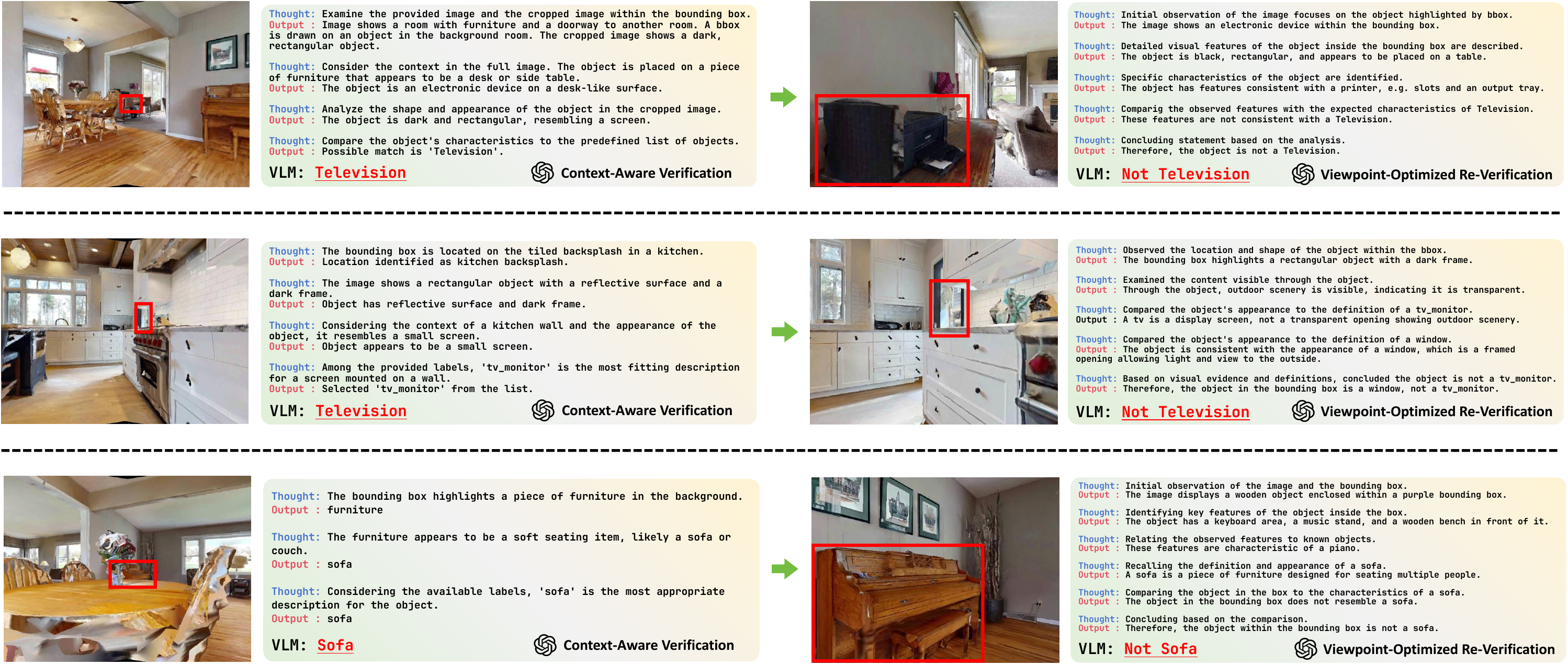}
    \caption{\textbf{Examples of viewpoint-optimized re-verification.}}
    \label{fig:re-verification_example}
\end{figure*}

We show more examples of viewpoint-optimized re-verification in \cref{fig:re-verification_example}. These examples demonstrate that we select a better viewpoint for re-verification, which can help the VLM to verify the target object more accurately. The selected viewpoint is not only closer to the target object but also provides a clearer view of the object, allowing the VLM to make a more confident verification decision.

    \section{Dataset Error}
\label{sec:dataset_error}
\begin{figure*}[h]
    \centering
    \includegraphics[width=0.95\linewidth]{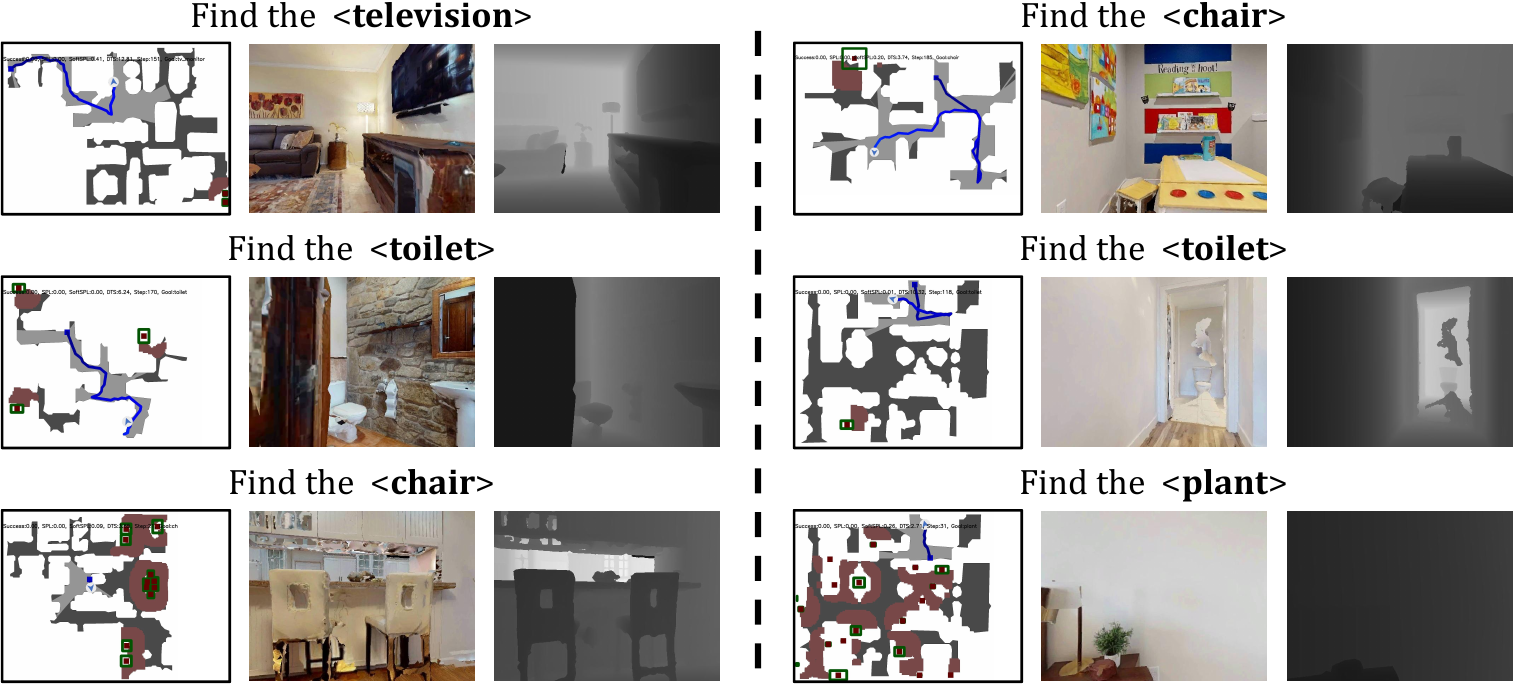}
    \caption{\textbf{Example of HM3D~\cite{ramakrishnan2021hm3d} dataset error causing episode failure.} The target object is not annotated in the scene, but the agent finds it. This episode will be incorrectly considered a failure.}
    \label{fig:hm3d_error}
\end{figure*}

\begin{figure*}[h]
    \centering
    \includegraphics[width=0.83\linewidth]{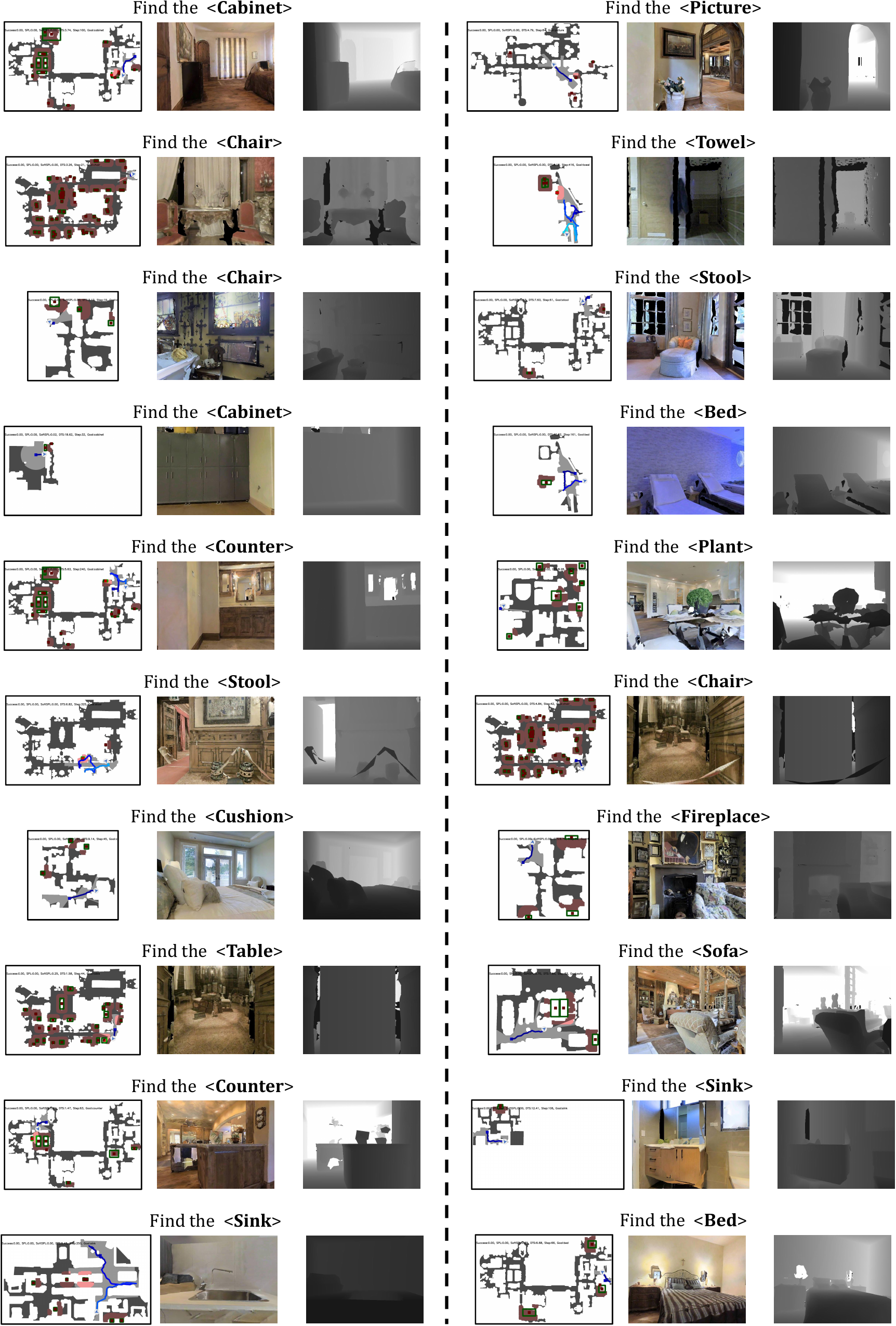}
    \caption{\textbf{Example of MP3D~\cite{Matterport3D} dataset error causing episode failure.} The target object is not annotated in the scene, but the agent finds it. This episode will be incorrectly considered a failure.}
    \label{fig:mp3d_error}
\end{figure*}

% 在这里，我提供了HM3D和MP3D数据集中的很多标注错误，主要是由于这两个数据集是从真实现实场景中采集的，并依赖人工标注其semantic信息，这导致了在很多场景中，遗漏了部分物体的标注，或者错误地标注了物体的semantic信息。在object navigation任务中，这导致的问题尤其严重，如果场景中漏标了targte object信息，但是agent实际上找到了那个漏标的targte object，你们这个episode会被错误的认为是失败的。我们在这里提供了HM3D和MP3D数据集中的一些标注错误的例子，供大家参考。
Here we provide many annotation errors in the HM3D~\cite{ramakrishnan2021hm3d} and MP3D~\cite{Matterport3D} datasets, mainly because these two datasets are collected from real-world scenes and rely on manual annotation of their semantic information. This leads to many scenes where some objects are missed or incorrectly annotated with semantic information. In the object navigation task, this problem is particularly serious. If the target object information is missing in the scene, but the agent actually finds that missing target object, this episode will be incorrectly considered a failure. We provide some examples of annotation errors in the HM3D and MP3D datasets for reference in \cref{fig:hm3d_error} and \ref{fig:mp3d_error}.
\vspace{20pt}

% \vspace{30pt}
% \clearpage

\clearpage

% \begin{figure}
%     \centering
%     \includegraphics[width=1.0\linewidth]{figures/dataset_error_mp3d_1.pdf}
%     \caption{Caption}
%     \label{fig:enter-label}
% \end{figure}

% \begin{figure}
%     \centering
%     \includegraphics[width=1.0\linewidth]{figures/dataset_error_mp3d_2.pdf}
%     \caption{Caption}
%     \label{fig:enter-label}
% \end{figure}
    \section{Room Segmentation}
\label{sec:room_seg}

Based on the scene point cloud, we first construct a top-down-view 2D histogram and extract the wall borders, resulting in a binary mask that highlights the walls in the scene. Next, we generate a whole-scene mask by combining the detected walls and point cloud. After obtaining the whole-scene mask, we gradually dilate the background to obtain the room segmentation.
% 具体来说，我们逐步膨胀墙壁，在每一步膨胀之后，我们会检查场景中的所有连通区域，如果某个区域的面积小于一个阈值，我们就将该区域标记为一个房间并从top-down-view图中移除这块区域。我们会一直膨胀墙壁，直到没有新的不联通区域被发现为止。最后，我们会将被标记的房间作为seed，执行watershed算法，得到最终的房间分割结果。
To be specific, we gradually dilate the background. After each dilation step, we check all the connected components in the scene. If a component's area is smaller than a threshold, we mark it as a room and remove that area from the dilated whole-scene mask. We continue dilating the background until no new disconnected regions are found. Finally, we use the marked rooms as seeds and apply the watershed algorithm on the whole-scene mask to obtain the final room segmentation results.

\begin{figure}[h]
    \centering
    \includegraphics[width=0.95\linewidth]{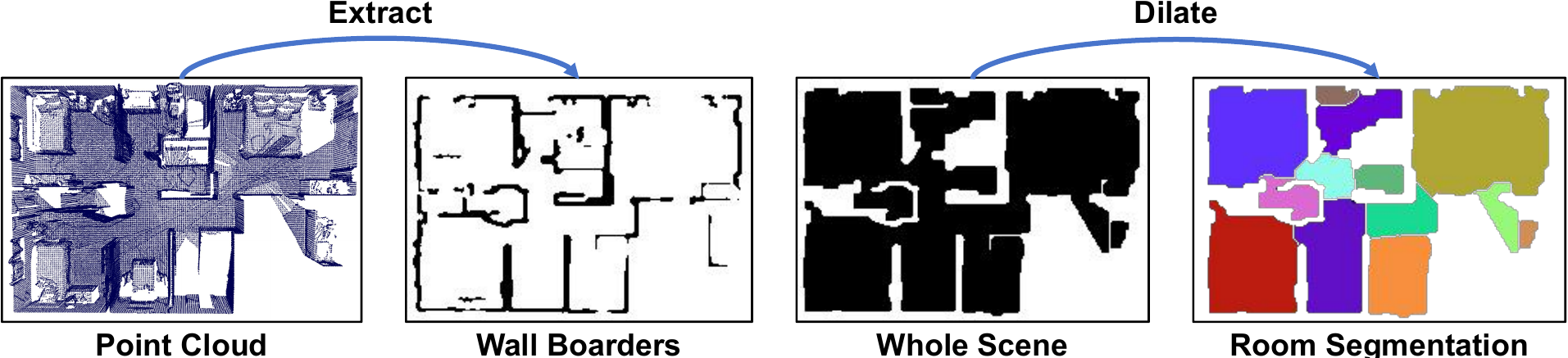}
    \caption{\textbf{Visualization of room segmentation process.}}
    \label{fig:enter-label}
\end{figure}

\end{document}